\documentclass[journal]{IEEEtran} 
\newcommand{\FIGPATH}{./fig/norm_res}

\usepackage{amsmath}  
\usepackage{amssymb}  
\usepackage{mathtools}

\usepackage{paralist}

\usepackage{verbatim}

\usepackage{color}
\usepackage{graphicx}
\usepackage{adjustbox}

\usepackage{mathptmx}
\usepackage{times}

\usepackage{fancyhdr}

\usepackage{multirow}

\newcommand{\neghphantom}[1]{\settowidth{\dimen0}{#1}\hspace*{-\dimen0}}

\usepackage[ruled,vlined,linesnumbered]{algorithm2e}
\usepackage[noend]{algorithmic}

\usepackage[us]{datetime}
\usepackage{caption}
\usepackage{subfloat}

\usepackage[color]{changebar_fix}
\usepackage{soul}
\newcommand{\rev}[1]{#1} 

\usepackage[normalem]{ulem}

\usepackage[usenames,dvipsnames,table]{xcolor}

\usepackage{hyperref} 
\hypersetup{
  colorlinks,
  citecolor=black,
  filecolor=black,
  linkcolor=black,
  urlcolor=black,
  pdfauthor={},
  pdfsubject={},
  pdftitle={}
}

\usepackage{todonotes}

\usepackage{subcaption}

\usepackage{latexsym}
\usepackage{color}

\usepackage{cite}

\usepackage[inline]{enumitem} 
\setenumerate[0]{label=\roman*.}

\usepackage{siunitx}
\usepackage{ifthen}

\usepackage{booktabs}
\usepackage{tablefootnote}

\usepackage{subfiles}
\usepackage{bm}

\setlist[itemize]{noitemsep, nosep}

\usepackage{pifont}
\definecolor{color_green}{rgb}{0, .722, .243}
\definecolor{color_red}{rgb}{0.737,0.165,0}
\newcommand{\greencheck}{{\color{color_green}\checkmark}}
\newcommand{\xmark}{{\color{color_red}{\ding{55}}}}

\newcommand\rot[1]{\rotatebox[origin=l]{60}{#1}}
\newcommand\objecttab[4]{\renewcommand{\arraystretch}{#1}\begin{tabular}{@{}l@{}}#3 \\\hphantom{vis}#4 \end{tabular}\renewcommand{\arraystretch}{#2}}

\usepackage{stackengine} 

\usepackage{tikz}
\usetikzlibrary{calc}
\usepackage{tikz-dimline}

\title{%
\Title
\\
{\small \SecondTitle}}

\usepackage[printonlyused]{acronym}
\acrodef{mav}[UAV]{Unmanned Aerial Vehicle}
\acrodef{ugv}[UGV]{Unmanned Ground Vehicle}
\acrodef{gps}[GPS]{Global Positioning System}
\acrodef{gnss}[GNSS]{global navigation satellite system}
\acrodef{mil}[MIL]{mirrorless interchangeable-lens}
\acrodef{sar}[S\&R]{Search and Rescue}
\acrodef{ctu}[CTU]{Czech Technical University in Prague}
\acrodef{pc}[PC]{personal computer}
\acrodef{lidar}[LiDAR]{light detection and ranging}
\acrodef{imu}[IMU]{inertial measurement unit}
\acrodef{rgb}[RGB]{color}
\acrodef{rgbd}[RGBD]{color-depth}
\acrodef{agl}[AGL]{above ground level}
\acrodef{amsl}[AMSL]{above mean sea level}
\acrodef{loam}[LOAM]{LiDAR Odometry and Mapping}
\acrodef{aloam}[A-LOAM]{Advanced implementation of LOAM}
\acrodef{liosam}[LIO-SAM]{LiDAR Inertial Odometry via Smoothing and Mapping}
\acrodef{dof}[DOF]{degrees of freedom}
\acrodef{lkf}[LKF]{linear Kalman filter}
\acrodef{mems}[MEMS]{micro-electromechanical systems}
\acrodef{fov}[FOV]{field of view}
\acrodef{slam}[SLAM]{simultaneous localization and mapping}
\acrodef{vio}[VIO]{visual-inertial odometry}
\acrodef{tio}[TIO]{thermal-inertial odometry}
\acrodef{lio}[LIO]{LiDAR-inertial odometry}
\acrodef{icp}[ICP]{Iterative Closest Point}
\acrodef{gicp}[GICP]{Generalized Iterative Closest Point}
\acrodef{ape}[APE]{Absolute Position Error}
\acrodef{ate}[ATE]{Absolute Trajectory Error}
\acrodef{gmm}[GMM]{Gaussian Mixture Models}
\acrodef{rssi}[RSSI]{Received Signal Strength Indicator}
\acrodef{rti}[RTI]{reflectance transformation imaging}
\acrodef{ptm}[PTM]{polynomial texture map}
\acrodef{uv}[UV]{ultraviolet}
\acrodef{ir}[IR]{infrared}
\acrodef{vis}[VIS]{visible spectrum photography}
\acrodef{vistr}[VISTR]{visible spectrum transmitography}
\acrodef{rak}[RAK]{raking light}
\acrodef{tpl}[TPL]{three point lighting}
\acrodef{vivl}[VIVL]{light-induced luminescence}
\acrodef{uvr}[UVR]{UV reflectography}
\acrodef{uvf}[UVF]{UV fluorescent photography}
\acrodef{uvrfc}[UVRFC]{false-color UV reflectography}
\acrodef{irr}[IRR]{IR reflectography}
\acrodef{irrtr}[IRRTR]{IR transmitography}
\acrodef{irf}[IRF]{IR fluorescent photography}
\acrodef{irrfc}[IRRFC]{false-color IR reflectography}
\acrodef{tsp}[TSP]{Traveling Salesman Problem}
\acrodef{ooi}[OoI]{object of interest}
\acrodef{mpc}[MPC]{model predictive control}

\newcommand{\Title}{New Era in Cultural Heritage Preservation:\\Cooperative Aerial Autonomy}
\newcommand{\SecondTitle}{Supervised Autonomy for Fast Digitalization of Difficult-to-Access Interiors of Historical Monuments}

\author{Pavel Petracek$^{*}_{\times}$, Vit Kratky$^{*}_{\times}$, Tomas Baca$^*$, Matej Petrlik$^*$, and Martin Saska$^*$
  \thanks{$^*$ Authors are with the Department of Cybernetics, Faculty of Electrical Engineering, Czech Technical University in Prague, Czech Republic.\\
          ${\times}$ \textit{Pavel Petracek and Vit Kratky are co-first authors.}\\
         {Corresponding author: \tt \scriptsize \href{mailto:pavel.petracek@fel.cvut.cz}{pavel.petracek@fel.cvut.cz}}%
  }
}

\begin{document}


\newcommand{\PREPRINTYEAR}{2023}
\newcommand{\PREPRINTPUBLISHER}{IEEE Robotics and Automation Magazine}

\fancyhead{}
\chead{\PREPRINTPUBLISHER, \PREPRINTYEAR. PREPRINT VERSION, DO NOT DISTRIBUTE.}
\pagestyle{fancy}
\thispagestyle{plain}

\onecolumn
\pagenumbering{gobble}
{
  \topskip0pt
  \vspace*{\fill}
  \centering
  \LARGE{%
    © \PREPRINTYEAR~\PREPRINTPUBLISHER\\\vspace{1cm}
    Personal use of this material is permitted.
    Permission from \PREPRINTPUBLISHER~must be obtained for all other uses, in any current or future media, including reprinting or republishing this material for advertising or promotional purposes, creating new collective works, for resale or redistribution to servers or lists, or reuse of any copyrighted component of this work in other works.\\\vspace*{1cm}DOI: 10.1109/MRA.2023.3244423}
    \vspace*{\fill}

}
\twocolumn 
\pagenumbering{arabic}
 
\maketitle

%



\begin{abstract}
  Digital documentation of large interiors of historical buildings is an exhausting task since most of the areas of interest are beyond typical human reach.
  We advocate the use of autonomous teams of multi-rotor \acp{mav} to speed up the documentation process by several orders of magnitude while allowing for a repeatable, accurate, and condition-independent solution capable of precise collision-free operation at great heights.
  The proposed multi-robot approach allows for performing tasks requiring dynamic scene illumination in large-scale real-world scenarios, a process previously applicable only in small-scale laboratory-like conditions.
  Extensive experimental analyses range from single-\acs{mav} imaging to specialized lighting techniques requiring accurate coordination of multiple \acs{mav}.
  The system’s robustness is demonstrated in more than two hundred autonomous flights in fifteen historical monuments requiring superior safety while lacking access to external localization.
  This unique experimental campaign, cooperated with restorers and conservators, brought numerous lessons transferable to other safety-critical robotic missions in documentation and inspection tasks.
\end{abstract}

\section{Autonomous Aerial Robotics for\\Heritage Digitalization}
\label{sec:introduction}
Digital documentation of large interiors of historical buildings is an exhausting task since most of the areas of interest are beyond typical human reach.
We advocate the use of fully-autonomous teams of cooperating multi-rotor \aclp{mav} (\acsp{mav}) to speed up the documentation process by several orders of magnitude while allowing for a repeatable, accurate, and condition-independent solution capable of precise collision-free operation at great heights.
In particular, we present a universal autonomy for \acp{mav} cooperating aerially within a team while documenting the interiors of historical buildings for the purposes of restoration planning and documentation works, as well as for assessing the structural state of aging historical sites.
We show that the proposed approach of active multi-robot cooperation enables performing documentation tasks requiring dynamic scene illumination in large-scale real-world scenarios, a process previously applicable only manually in areas easily accessible by humans.

\begin{figure*}
  \centering

  \input{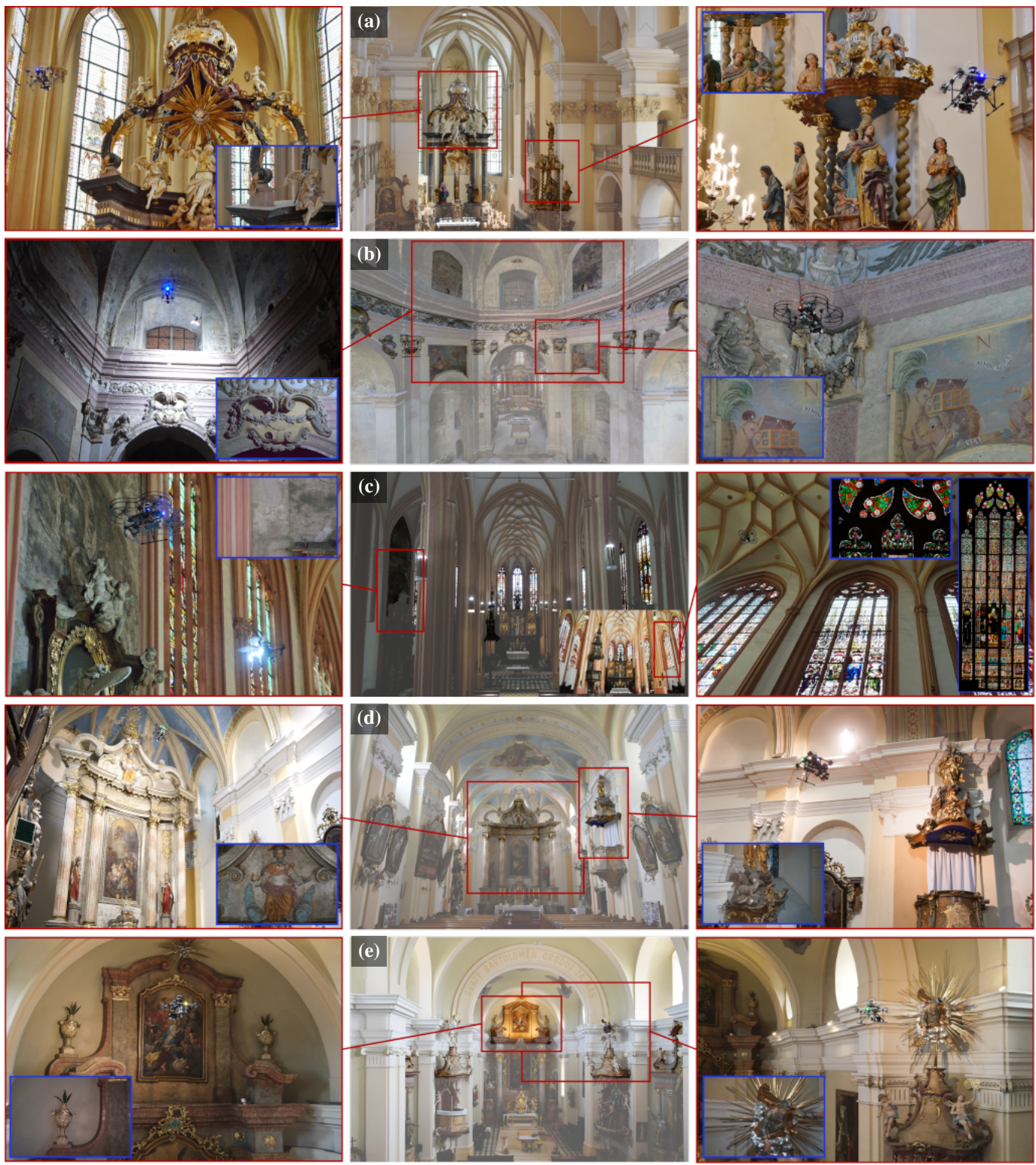}
  \vspace{-1.2em}
  \caption{Illustration of deployment of the presented methodology in selected historical buildings located in the Czech Republic ---
  (a) Church of the Exaltation of the Holy Cross in Prost\v{e}jov,
  (b) St. Anne and St. Jacob the Great Church in Star\'{a} Voda by Libav\'{a},
  (c) Church of St. Maurice in Olomouc,
  (d) Church of the Nativity of the Virgin Mary in Nov\'{y} Mal\'{i}n, and
  (e) Church of St. Bartholomew in Z\'{a}b\v{r}eh.
  Center images show the interiors of the churches with highlighted objects of documentation interest.
  Side images show actual deployment of \acp{mav} in the particular settings together with example images (highlighted in blue) captured by an onboard camera.%
  }
  \label{fig:deployment_mosaic}

\end{figure*}

The presented system was developed in cooperation with cultural heritage institutions as part of the Dronument project~\cite{mrs_dronument_page} and was deployed fully autonomously in numerous characteristically diverse historical monuments, as exhibited in~\autoref{fig:deployment_mosaic} and~\autoref{tab:documentation_summary}.
The included experimental evaluation utilizes \acp{mav} in multiple real-world documentation tasks, and discusses the quality of the obtained results used in subsequent restoration works, as well as suitability of particular techniques for \acp{mav}.  
The analyses demonstrate the framework's robustness in single and multi-robot deployments in more than two hundred fully-autonomous flights in fifteen historical monuments.
In these experiments, the aerial robots rely solely on onboard sensors without access to external localization such as \acp{gnss} or motion capture systems, which significantly increases deployability of the system.
This unique, extensive, experimental campaign, which cooperated with restorers and conservators, brought numerous lessons learned that are transferable to other safety-critical robotic missions in documentation and inspection tasks.
The system also serves as a large part of an official methodological study approved by the Czech National Heritage Institute for its high added value in heritage protection.
The methodology (available at~\cite{mrs_dronument_page}) describes the proper usage of \acp{mav} in historical structures for the first time and so prescribes the proposed system to be a standard in this application.




\section{Background}
Often serving educational, cultural, or social purpose, the preservation of cultural heritage as a valuable reminder of our history is in the greater interest of society. 
Cultural management and preservation of historical monuments became a relevant topic in the late $19^{\text{th}}$ and $20^{\text{th}}$ centuries when many valuable historical monuments were destroyed while establishing modern infrastructure.
By introducing cultural heritage preservation into legislation, the monuments gained protection from human interference.
However, being exposed to real-world conditions continually degrades historical buildings and artifacts within.
This has initiated the endeavor to actively prevent the irreversible damage of cultural heritage by monitoring its condition and performing restoration and conservation works.

Conservation work on a historical artifact comprises four consecutive phases: the initial survey, the choice of restoration steps and costs evaluation, the actual restoration works, and continued monitoring of the restoration.
Both the initial survey and monitoring phase require providing information about the artifact in digital form (usually camera imaging). 
Thus, these phases are considered a data collection task for which an aerial vehicle, capable of gathering data in a cost-effective and fast manner, can be of great help. 
This is especially true for areas of interest which are located beyond typical human reach, a situation often arising in tall historical buildings such as churches and cathedrals.
Apart from planning restoration works, gathered digital materials can support the reconstruction of a structure in the event of its sudden accidental destruction (e.g., the burning of the Notre-Dame Cathedral in 2019).




\section{Robotics and Automation in Cultural Heritage Preservation}
Documentation and digitalization of historical objects requires gathering various types of data, e.g., camera images in visible, \ac{ir} and \ac{uv} spectra, and 3D models. 
The data gathering is demanding in both time and human resources, particularly in large buildings.
This motivates the endeavor to automate data gathering by introducing mobile robotic solutions capable of fast autonomous documentation.
The first level of mobile-robot automation can be achieved by applying \acp{ugv} as carriers of the documentation sensors.
A \ac{ugv} equipped with a laser scanner and capable of autonomous navigation in constrained environments can sequentially visit several locations to collect a set of scans covering the entire operational space~\cite{prieto2017ras}.
An advantage of this approach lies primarily in reducing necessary human participation in the scanning process, allowing for the collection of scans from potentially dangerous areas.
Several systems applying such an approach were already developed and deployed for scanning historical monuments~\cite{blaer2007ugvdocumentation, borrmann2015ugvmapping}.

Whereas the operational space of \acp{ugv} usually does not exceed typical human reach, multi-rotor \acp{mav} capable of 3D navigation in confined environments can be applied for data collection tasks in difficult-to-access areas.
In exteriors, \ac{mav} solutions abundantly utilize predefined \ac{gnss} poses for navigation~\cite{bakirman2020culturaldocumentation}.
In contrast to exteriors, the applicability of \acp{mav} in interiors imposes additional challenges --- lack of \acs{gnss} localization, navigation in a confined environment, and non-negligible aerodynamic effects.
Because of that, \ac{mav} systems deployed for indoor data gathering are mainly limited to industrial inspections, with only a few works targeting \acs{mav}-based documentation of historical buildings.
The specifics of such an application are targeted in this work.

For industrial inspections, the literature typically exploits the environment structure, such as known profiles of tunnels~\cite{ozaslan2017inspection} or structured and well-lit warehouses~\cite{beul2018warehouses}.
More general solutions were introduced in the commercial sector introducing semi-autonomous \ac{mav} inspection systems\footnotemark --- DJI Mavic~3, Elios~3, or Skydio~2+\textsuperscript{\texttrademark}.
In interiors, DJI provides image-based \ac{mav} stabilization, Elios allows for human-operated flight with \acs{lidar} and camera-based stabilization and mapping with guarantees of environmental and mechanical protection, and Skydio\textsuperscript{\texttrademark} offers automated camera-stabilized flight for interactive 3D reconstruction.
Although all these solutions provide an assistive level of autonomy in \ac{mav} stabilization, the first two require human-in-the-loop navigation.
None of the mentioned solutions offer full interior autonomy, repeatability, modularity, rotor nor sensory redundancy, imaging focusing on capturing high-quality details, and cooperative multi-robot deployment.
\footnotetext{%
  DJI Mavic~3: \href{https://www.dji.com/cz/mavic-3}{dji.com/cz/mavic-3},
  Elios~3: \href{https://www.flyability.com/elios-3}{flyability.com/elios-3},
  Skydio~2+\textsuperscript{\texttrademark}: \href{https://www.skydio.com/skydio-2-plus}{skydio.com/skydio-2-plus}.}

As mentioned, aerial data gathering inside historical buildings is rare.
A specialized platform for assisting in cultural heritage monitoring called \textit{HeritageBot} was introduced in~\cite{ceccarelli2017heritagebot}.
However, no evidence of the deployment of this platform in historical monuments is presented.  
In~\cite{hallermann2015}, the authors propose an assistive system to manual control of the \ac{mav} during inspection tasks with the experimental deployment of the system inside and outside historical sites.


Among introduced solutions, the most advanced \acs{mav}-based systems with the high level of autonomy required for the interiors of historical buildings were introduced in our recent works~\cite{petracek2020ral,kratky2020ral,smrcka2021icuas,kratky2021documentation}.
In these publications, we introduced
a preliminary application-tailored autonomous \ac{mav} system allowing for safe localization and navigation inside historical structures~\cite{petracek2020ral},
the methodology and algorithms for the realization of advanced documentation techniques found in \ac{rti}~\cite{kratky2020ral} and \ac{rak}~\cite{smrcka2021icuas}, and
an autonomous single-\acs{mav} system for realization of documentation missions~\cite{kratky2021documentation}.
All works provide a fully autonomous solution and the possibility of performing documentation techniques in difficult-to-access areas without using mobile lift platforms or scaffolding installation.
Here, we progress beyond previous works by introducing a full 3D \ac{slam} methodology for indoor localization of robots;
by advancing robustness to localization drifts and hard-to-detect obstacles with additional sensory redundancy;
by improving path, trajectory, and mission planning;
by using a \ac{mav} team to realize documentation techniques that could not be realized with only a single robot in principle; and
by presenting the complete set of results achieved in the Dronument project that are summarized in numerous lessons learned during the unique experimental campaign within highly safety-critical missions.







\section{Documentation Techniques and Associated Constraints}\label{sec:documentation_techniques}

\begin{table*}
\renewcommand{\arraystretch}{1.2}
  \begin{center}
    \newcommand{\rotateAngle}{00}
    \newcommand*{\OK}{\greencheck}
    \newcommand*{\cOK}{{\ding{113}\raisebox{.30ex}{\hspace{-0.75em}$\color{color_green}\checkmark$}}}
    \caption{Recapitulative table of documentation tasks selected as realizable by aerial vehicles in interiors of historical structures.
             The squared check marks $\left(\text{\cOK}\right)$ identify the realizable documentation methods which were experimentally applied in historical structures, as summarized in \autoref{sec:exp}.
             The last column marks methods for which the ambient light is either required (\OK), forbidden (\xmark), or arbitrary (unmarked).  
    }\label{tab:documentation_tasks}
    \begin{tabular}{l l  c c c c  c}
      \toprule 
      ~ & ~ & \multicolumn{2}{c}{Realizable by} & \multicolumn{3}{c}{Required equipment and lighting conditions}\\\cmidrule(lr){3-4}\cmidrule(lr){5-7}
      ~ & {\hfill Documentation technique} & \rotatebox{\rotateAngle}{Single robot} & \rotatebox{\rotateAngle}{Multiple robots} & \rotatebox{\rotateAngle}{Onboard camera} & \rotatebox{\rotateAngle}{Onboard light} & \rotatebox{\rotateAngle}{Ambient light}	\\
      \midrule 
                              & visible spectrum: photography (\acs{vis})			                              & \cOK & ~ & \OK	& \OK & \OK	\\  
                              & \hphantom{visible spectrum:} transmitography (\acs{vistr})                	& ~	& \OK & \OK	& \OK & ~	\\  
                              & \hphantom{visible spectrum:} raking light (\acs{rak})			                  & \cOK	& \OK & \OK	& \OK & ~\\  
                              & \hphantom{visible spectrum:} three point lighting (\acs{tpl})			                  & ~	& \cOK & \OK	& \OK & ~	\\  
                              & \hphantom{visible spectrum:} \acl{rti} (\ac{rti})	& \cOK	& \cOK & \OK	& \OK & \xmark\\  
                              & \hphantom{visible spectrum:} light-induced luminescence (\acs{vivl})			  & \OK	& ~ & ~	& \OK & ~\\  
                              & \acs{uv} spectrum:\hphantom{visible}\neghphantom{\acs{uv}} reflectography (\acs{uvr})	  & \cOK	& ~   & ~	& \OK & \xmark \\  
                              & \hphantom{visible spectrum:} fluorescent photography (\acs{uvf})			      & \cOK	& \OK & \OK	& \OK & ~\\
                              & \hphantom{visible spectrum:} false-color reflectography (\acs{uvrfc})			  & \OK	& ~ & ~	& \OK & \xmark\\  
                              & \acs{ir} spectrum:\hphantom{visible}\neghphantom{\acs{ir}} reflectography (\acs{irr})		& \cOK	& ~ & ~	& \OK	& \xmark\\ 
                              & \hphantom{visible spectrum:} transmitography (\acs{irrtr})			            & \OK	& ~ & ~	& \OK & \xmark\\  
      \multirow{-12}{*}{\rotatebox{90}{Spectral analysis}} 
                              & \hphantom{visible spectrum:} fluorescent photography (\acs{irf})			      & \cOK	& \OK & \OK	& \OK & ~\\  
                              & \hphantom{visible spectrum:} false-color reflectography (\acs{irrfc})		    & \OK	& ~ & ~	& \OK & \xmark\\  
                              & X-ray:\hphantom{visible spectrum:}\neghphantom{X-ray:} radiography		& ~	& \OK & ~	& ~	& ~\\ 
      \midrule
                              & 3D reconstruction	        & \cOK	& \OK & \OK	& ~ & ~\\
                              & photogrammetry		        & \OK & \OK & \OK	& ~ & \OK \\
      \multirow{-3}{*}{\rotatebox{90}{Others}}                       
                              & environmental monitoring	& \OK & \OK & ~	& ~ & ~ \\
      \bottomrule
    \end{tabular}
  \end{center}
\end{table*}

The documentation techniques applied in the field of restoration and cultural heritage preservation aim to capture the current state of the object, survey a potential structural or artistic damage, and determine the age, author and possible dimensions of the elements by identifying the materials and techniques that have been used.
For this purpose, diverse methods combining conventional photography in the visible spectrum, photography in invisible spectra making use of different reflective properties of materials, specialized lighting techniques applied for revealing structural details, and even invasive methods based on the collection of material samples are applied.
In robotic context, all these methods are associated with varying requirements on sensory equipment, amount of cooperation, and external conditions (mainly illuminance).
These relations are summarized in \autoref{tab:documentation_tasks}, together with the studied documentation techniques.

The most common documentation technique providing initial information about the studied subject is standard \ac{vis}.
This technique is applicable to all types of studied objects, ranging from flat paintings and frescoes to 3D structures, including statues and altars.
Since the documented areas of historical buildings are often dark, the obtained images suffer from insufficient lighting conditions.     
Hence, the \ac{vis} method often requires additional external lighting to locally increase illuminance, allowing for the decreased exposure times required to avoid motion blur \mbox{from instabilities of a multi-rotor vehicle.}

Similar to aesthetic photography, light plays a significant role in restoration documentation.
Documentation techniques capturing data in the visible spectrum make use of varying lighting intensity and illumination angles to enhance the quality and amount of information that can be derived from the gathered data.
The main group of lighting techniques applicable during documentation tasks aims to highlight the 3D characteristics of captured objects, with \ac{tpl} being the most routine.
\ac{tpl} illuminates the object with several sources of luminance, each with different intensity and orientation with respect to the camera's optical axis, in order to provide an aesthetically pleasant and realistic view of the 3D object.
Another widely used lighting technique is \acl{rak} (\acs{rak}), which focuses on revealing the surface details of flat objects. 
While \ac{tpl} employs several light sources to avoid overshadowed areas, \ac{rak} applies a single light as parallel to the scene as possible.
The illumination angle in \ac{rak} exploits the shadows to highlight the roughness of the surface.

A highly specialized documentation technique used in the field of restoration is the \acl{rti} (\acs{rti}) --- an image-based rendering method used for obtaining a representation of an image that enables displaying the image under an arbitrary direction of illumination.
The necessary inputs of this method include a set of images of an object taken by a static camera, with each image being under illumination from a different but known direction.
The captured images and the corresponding lighting vectors are then used for the computation of a \ac{ptm} representation of the image that enables an interactive illumination and view of the object.
Another specialized documentation technique is \ac{vistr} which requires a light source to be positioned behind an \ac{ooi} to transmit the light through this object.
However, this method is mainly applied for canvas paintings and thus is rather impractical for realization by \acp{mav}.
 
Multiple techniques exploit \ac{uv} and \ac{ir} lumination and its effects.
While the methods based on the visible light focus on revealing structural characteristics and colors, the \ac{uv} and \ac{ir} methods aim primarily to identify the materials and hidden layers of artworks.
The use of different spectra allows more precise dating of the paintings, as the glow of pigment combinations are unique to certain periods.
The first group of methods applying \ac{uv} and \ac{ir} lights is based on capturing the fluorescent light in the visible spectrum emitted by an object after absorbing \ac{uv} or \ac{ir} radiation energy.  
These methods are called \ac{uvf} and \ac{irf} and are used for, e.g., detecting zinc and titanium white (\acs{uvf}) or cadmium red and Egyptian blue (\acs{irf}).  
The second group of methods applying \ac{uv} and \ac{ir} lights captures the reflected light in the corresponding spectra.
These methods are called \ac{uvr} and \ac{irr} and are applicable for, e.g., detecting restored areas, highlighting repairs and re-touchings, enhancing faded paintings (\ac{uvr}) or \mbox{reaching the underdrawing layer of paintings (\ac{irr}).}
  
Except for \ac{vis}, all the above-mentioned methods require positioning the light at a certain angle with respect to the camera.
Therefore, these methods are not fully realizable by a single \acs{mav} and require a multi-robot coordination.
The particular methods can be realized in three different configurations dependent on the requirements of the task.
The first configuration employs an autonomous multi-robot team consisting of a \ac{mav} carrying a documentation sensor and a set of supporting \acp{mav} providing dynamic lighting of the documented scene.
The second configuration applies the \ac{mav} as a carrier of the sensor whilst the light is provided by external sources.
The third configuration uses the \ac{mav} for positioning the light whereas the data are captured by a static sensor from the ground.

The largest problem in realization of the techniques relying on a \ac{mav} carrying a camera is the exposure time required for sharp and detailed imaging.
\autoref{tab:exposure_times} summarizes that the exposure times for some of the methods reach tens of seconds.
With constraints on image sharpness, such long times and natural \textit{nonstaticity} of highly dynamical multi-rotor \acp{mav} prevent the realization of these techniques in the camera-carrier mode with satisfactory results.
Instead, imaging with a static camera and aerial lighting was investigated for some of these techniques.

\begin{table}
\renewcommand{\arraystretch}{1.2}
  \begin{center}
    \caption{Typical exposure times of the selected documentation techniques.%
    }\label{tab:exposure_times}
    \newcommand{\rotateAngle}{00}
    \newcommand*{\OK}{\checkmark}
    \newcommand*{\hshift}{\hspace*{0.0cm}}
    \begin{tabular}{l c r}
      \toprule 
      Technique & Spectrum & {Exposure time (s)}\\
      \midrule
      \acl{vis}   & visible & $\leq0.2$\hshift \\
      \acl{rak}   & visible & $\leq0.2$\hshift \\
      \acl{tpl}   & visible & $\leq0.2$\hshift \\
      \acl{rti}   & visible & $\leq0.2$\hshift \\
      \acl{uvf}   & \acs{uv} & $\leq2.0$\hshift \\
      \acl{uvr}   & \acs{uv} & 2.0\hshift \\
      \acl{vistr} & visible & 2.0\hshift \\
      \acl{irr}   & \acs{ir} & 4.0\hshift \\
      \acl{irrtr} & \acs{ir} & 20.0\hshift \\
      \acl{vivl}  & visible & 25.0\hshift \\
      \acl{irf}   & \acs{ir} & 30.0\hshift \\
      radiography & X-ray & $\geq30.0$\hshift \\
      \bottomrule
    \end{tabular}
  \end{center}
\end{table}





The non-spectral tasks applied in the field of preservation mostly focus on the 3D reconstruction and environment monitoring through static sensors measuring physical quantities (e.g., temperature, humidity).
The most common techniques applied in 3D reconstruction use visible spectrum images (photogrammetry) or scans produced by laser sensors.
From the perspective of the proposed system, the data gathering process for 3D reconstruction does not differ from the realization of \ac{vis} and collection of raw data from onboard sensors used for localization and mapping.
Monitoring the physical quantities in an environment requires attaching a sensor to the \ac{mav} frame and navigating it to the required area.
If the measurement process requires permanent monitoring, the sensor must be attached at a specific position in the environment (e.g., adhered to a wall or placed on a mantel).
This process is also realizable by \acp{mav} but requires fine control, state estimation, and a mechanism for physical robot-to-environment interaction, as closely tackled in~\cite{smrcka2021icuas}.

\section{\acs{mav}-based Framework for Documentation of Cultural Heritage Interiors}\label{sec:archi}

\begin{figure*}
  \begin{center}
    \input{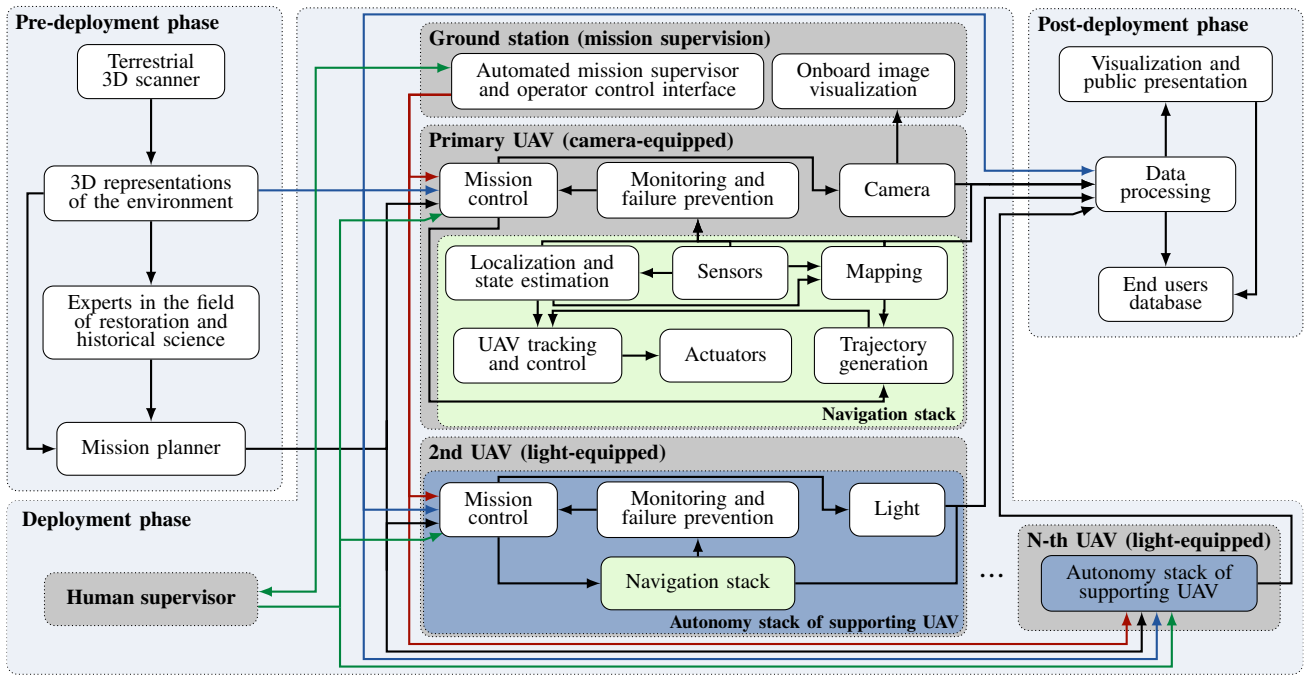}
    \caption{High-level diagram of the three-phase architecture of the system designed for multi-\acs{mav} documentation of interiors of historical buildings.
    The 3D model of the environment and the mission plan are used as an a-priori generated input for the realization of the documentation mission itself.
    After the deployment phase, the data gathered during the mission are processed and provided to the end users.} 
    \label{fig:sys_archi} 
  \end{center}
\end{figure*}

The overall pipeline of the \acs{mav}-based framework for interior documentation in historical monuments is showcased in~\autoref{fig:sys_archi}.
The framework is composed of three main phases --- the pre-deployment phase incorporating pre-flight data gathering and mission planning, the actual deployment of the system in interiors of historical buildings, and post-deployment phase, including processing and utilization of the collected data. 

\subsection{Pre-deployment Phase}\label{sec:predeployment}

\begin{figure*}
  \centering

  \input{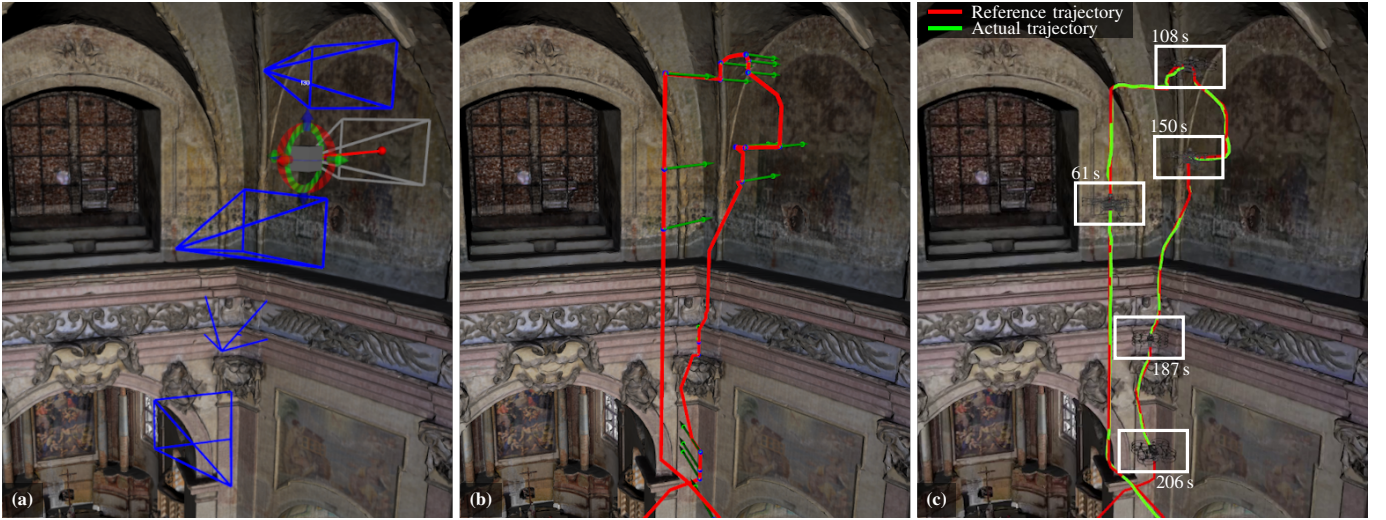}
  \vspace{-1.2em}
  \caption{Pre-deployment phase of the proposed framework --- (a) specification of the documentation task by selecting a set of camera viewpoints within the 3D model of the environment, (b) planning trajectory of the robot (in red) which visits all the specified viewpoints (in green), and (c) verification of the mission plan in Gazebo simulator employing identical software that is used during real-world missions.}
  \label{fig:pre_deployment_phase}

\end{figure*}

The first step preceding the entire documentation process is obtaining a model of the environment used for safe navigation of the \ac{mav}, as well as for the specification of \acsp{ooi} that should be scanned during documentation missions.
For this purpose, a precise terrestrial 3D scanner Leica BLK360 is employed to obtain a set of scans that are later used for building a complete 3D representation of the target environment, both in form of a global point cloud and a 3D model with a colored texture.
The colored 3D model serves for precise specification of the desired camera viewpoints and for presenting the documentation outputs to the public and the end users.
The camera viewpoints specifications are made by experts of restoration or historical science who position a virtual camera within the 3D model of the environment using \rev{a viewpoint-selection tool shown in~\mbox{\autoref{fig:pre_deployment_phase}a}.
This tool shows a camera and its view and enables to save the camera viewpoint pose in the global coordinate frame.
The optical properties of the camera} can be parameterized with respect to the equipment available for real-world documentation, thus allowing for visualizing the desired photo to be captured from a given pose in the colored 3D model.



Given the point cloud representation of the environment and the set of to-be-captured images represented by the respective camera viewpoints \rev{in the global coordinate frame}, the documentation mission plan is generated as follows. 
First, the problem of finding an optimal sequence $\sigma^*$ of camera poses minimizing the overall traveled distance is defined and solved as the \ac{tsp}.
Considering the possible dimensionality of the problem, a solver using an efficient Lin-Kernighan heuristics~\cite{lkh_algorithm} is employed for the solution of \ac{tsp} to enable on-site plan generation. 
Constrained by available computational time, the mutual distance between particular pairs of poses within the solution of \ac{tsp} are given either by the Euclidean distance or by the length of the collision-free path between the poses.
Second, the consequent poses in $\sigma^*$ are connected by the collision-free paths generated with the use of a grid-based planner~\cite{kratky2021exploration}. 
This process creates a path connecting all the poses which can be generally unfeasible if limited flight time of a \ac{mav} is taken into account.
Hence, the final set of plans $\mathbb{P} = \{\mathbf{P}_1,\mathbf{P}_2, \ldots, \mathbf{P}_n\}$ is obtained by splitting $\sigma^*$ to a set of subsequences $\mathrm{\Sigma} = \{\mathbf{\sigma}_1, \mathbf{\sigma}_2, \ldots, \mathbf{\sigma}_n\}$,
where \mbox{$\mathbf{\sigma}^* = \mathbf{\sigma}_1 \cup \mathbf{\sigma}_2 \cup \ldots \cup \mathbf{\sigma}_n$},
$\mathbf{P}_i \in \mathbb{P}$ is a collision free path connecting the initial pose with a sequence of poses in $\mathbf{\sigma}_i$, and equation
$\text{t}(\mathbf{P}_i) < t_{max}$ holds, $\forall i \in \{1, \ldots, n\}$, for $t_{max}$ being the maximum flight time of the \ac{mav}, and $\text{t}(\mathbf{P}_i)$ being the time needed for following path $\mathbf{P}_i$.


To increase the mission safety, the final step of the pre-deployment phase verifies the paths planned for the documentation mission.
First, each plan is verified by humans as collision-free by visualizing it in the 3D model of the environment.
Second, the plan feasibility is verified by simulating the entire mission in the realistic Gazebo simulator using the virtual model of the environment with the same software and sensory plugins used during real-world missions.
The goal of this two-stage process is to verify that all the generated paths are collision-free and do not traverse potentially risky parts of the environment.
The mission specification and plan validation is showcased in~\autoref{fig:pre_deployment_phase}.


\subsection{\ac{mav} Deployment}\label{sec:deployment}

The actual system deployment is influenced by the application's specificity imposing strict safety-guarantee requirements.
After the necessary hardware checks, all software components are initialized on the onboard computer of each \ac{mav}.
After successful initialization, all the \acp{mav} automatically align their reference frame with the common frame of coordination by matching their sensory data to the sparse interior map available to each \ac{mav}.
The outputs of this phase are visually verified by the operator, who checks the correctness of the frames' alignment and validates the mission plan for the last time.

During the following autonomous mission, an automatic centralized supervisor (a ground station) checks the state of all the \acp{mav} in real-time.
This supervisor reacts to faults and allows for revealing many possible failures, even preventatively.
Available safety actions include stopping all the airborne \acp{mav} at the place at once, navigating them cooperatively to takeoff locations, and landing them at safe locations.
Apart from the automatic supervisor, all these actions can be triggered by a human operator supervising the mission in parallel using the ground station.
At last, a human operator serves as the final safety measure capable of landing the \acp{mav} manually. 
The autonomy stack is described in~\autoref{sec:system_architecture}.

\subsection{Post-deployment Phase}\label{sec:postdeployment}

To increase the quality and range of the outputs, the data collected during autonomous flights in historical buildings are processed before being provided to the end users.
This includes post-processing of onboard sensory data to increase accuracy of pose referencing associated with the captured data frames, stitching images into photomaps, or building a 3D model of the environment in areas occluded in ground-located scans.
The generated data then serve for digitalization and archivation, pre- and post-restoration analyses, state assessment and monitoring, material analyses, photogrammetry, and for digital presentation to the public.



\section{Fully Autonomous, Cooperating \acp{mav}}
\label{sec:system_architecture}


To benefit from extensively tested and field-verified methods, the proposed multi-\acs{mav} system is based on the open source MRS UAV system\protect\footnotemark~developed within the authors' research group.
In this section, let us summarize novel scientific results achieved within the presented project Dronument, whilst the MRS UAV system is described in detail in~\cite{baca2021mrs}.

\footnotetext{\href{https://github.com/ctu-mrs/mrs_uav_system}{github.com/ctu-mrs/mrs\_uav\_system}}


\subsection{Reference Frame Alignment}\label{sec:frame_alignment}

The reference frames of the robots are aligned once during a pre-takeoff phase with each robot performing the alignment independently in four automated phases.
This alignment process is mandatory for each robot as the supervising controller does not allow any robot to takeoff unless all robot frames are aligned with the global coordination frame (i.e., the map).

In the \textit{data loading} phase, each robot loads the global map $\mathbf{M}$ and a single 3D \acs{lidar} data-frame $\mathbf{D}$ to its memory, applies voxelization to both the objects for dimensionality reduction, and removes outliers in $\mathbf{D}$ using radius outlier filter.
The z-axis of both the point clouds $\mathbf{M}$ and $\mathbf{D}$ is assumed to be approximately parallel to the gravity vector.
During \textit{global correlation} phase, the origins and orientations of $\mathbf{M}$ and $\mathbf{D}$ are approximately matched.
First, convex 3D-space hulls $\mathbf{H_M}$ and $\mathbf{H_D}$ are computed using Qhull~\cite{qhull} with a hull being represented as a set of undirected edges $\mathbf{H} = \left\{\left(\mathbf{v}_a, \mathbf{v}_b\right)_i\right\}$ (set of vertex pairs).
Translation $\mathbf{t_D^M} \in \mathbb{R}^3$ of $\mathbf{D}$ to $\mathbf{M}$ is given as $\mathbf{t_D^M} = \mathbf{b_M} - \mathbf{b_D}$, where $\mathbf{b_{X}} \in \mathbb{R}^3$, $\mathbf{X \in \left\{M, D\right\}}$, represents a polyline barycenter of an edge set $\mathbf{X}$ as
\begin{equation}
  \mathbf{b_{X}} = \dfrac{\sum_{\left(\mathbf{v}_a, \mathbf{v}_b\right) \in \mathbf{X}} \left[\mathbf{v}_a + \left(\mathbf{v}_b - \mathbf{v}_a\right)/2\right] ||\mathbf{v}_b - \mathbf{v}_a||_2 }{\sum_{\left(\mathbf{v}_a, \mathbf{v}_b\right) \in \mathbf{X}} ||\mathbf{v}_b - \mathbf{v}_a||_2}.
\end{equation}
The \ac{mav} is assumed to be taking off from ground locations, hence the grounds are coupled by setting z-axis translation to $\mathbf{t_D^M}(z) = \min_{\mathbf{p} \in \mathbf{M}}\mathbf{p}(z) - \min_{\mathbf{p} \in \mathbf{D}}\mathbf{p}(z)$\rev{, where $\mathbf{p}(z)$ denotes the $z$ coordinate of point $\mathbf{p}$}.
Initial transformation to the consequent optimization phases is then given as 
\begin{equation}
  \mathbf{T_I} = \mathbf{T(t_D^M)} \mathbf{T}\left(\mathbf{t_D}, \mathbf{z}, \theta\right),
\end{equation}
where $\mathbf{T}~\in~\mathbb{R}^{4 \times 4}$ is a general 3D transformation in the matrix form and
$\mathbf{T}(\mathbf{t_D}, \mathbf{z}, \theta)$ is the matrix form of a z-axis rotation at a point $\mathbf{t_D} \in \mathbb{R}^3$ (the origin of $\mathbf{D}$) by angle $\theta$.
The rotation angle is given as $\theta = \theta_{\mathbf{M}} - \theta_{\mathbf{D}}$, where \rev{$\theta_{\mathbf{X}}~=~\arctan{\xi^{\mathbf{X}}_y/\xi^{\mathbf{X}}_x}$,
$\mathbf{\xi}^{\mathbf{X}} = \left(\xi^{\mathbf{X}}_x, \xi^{\mathbf{X}}_y, \xi^{\mathbf{X}}_z\right) = \arg\max_{\mathbf{\xi} \in \Xi(\mathbf{X})}\sqrt{\xi_x^2 + \xi_y^2}$}, and
$\Xi(\mathbf{X})$ is the set of covariance matrix eigenvectors of the point cloud~$\mathbf{X}$.

The following \textit{global registration} phase copes with the lateral symmetry of the environments as typical of large historical structures.
Several \ac{icp} routines $\text{ICP}(\mathbf{T})$ are performed in this phase, each with different initializations $\mathbf{T}$ and loosely set parameters for point association and convergence requirements.
Given a number of desired initializations $k$, this phase selects $\theta^*~=~\arg\min_{\theta \in \Theta}~\text{ICP}\left(\mathbf{T_I}\mathbf{T}(\mathbf{t_D}, \mathbf{z}, \theta)\right)$ where \mbox{$\Theta = \left\{2\pi i/k~|~i \in \left\{0, 1, \ldots, k-1\right\}\right\}$.}
Final \textit{fine-tuning optimization} phase estimates robot origin in the global coordinate frame $\mathbf{T_D^M}$ by running $\text{ICP}\left(\mathbf{T_I}\mathbf{T}(\mathbf{t_D}, \mathbf{z},\theta^*)\right)$ optimization set with high-accuracy parameters and strict convergence criteria.

\subsection{State Estimation, Localization, and Mapping}\label{sec:estimation_localization_mapping}

Estimating the 3D state of a \ac{mav} (i.e., pose and its derivatives) in real-time is crucial for the \ac{mav} mid-air control and 3D navigation.
To keep the robot steady while airborne, follow reference trajectories, and avoid obstacles, the environment needs to be perceived with robot's onboard sensors (e.g., cameras, \acsp{lidar}).
As state estimation, localization, and mapping are critical for collision-free flight, the utilized algorithms are based on well-tested works with implementation validated in differing real-world scenarios.
To estimate the robot state, a bank of Kalman filters~\cite{baca2021mrs} extended with smoothing over a short past-measurements buffer fuses onboard inertial measurements with localization outputs, providing real-time feedback to the position control loop~\cite{baca2021mrs}.
\rev{%
The localization and mapping systems utilize low-drift pose estimation LOAM\mbox{~\cite{zhang2017loam}}.
An extensive evaluation in\mbox{~\cite{kratky2021exploration}} showed that fusing LOAM efficiently with\mbox{~\cite{baca2021mrs}} provides sufficient accuracy and robustness} even in safety-critical applications.
The architecture of the control, state estimation, and localization pipelines is analogous to~\cite{kratky2021exploration}.
In contrast to~\cite{kratky2021exploration}, the mapping pipeline uses an a-priori map of the environment to derive a global frame for the robots' missions (see its calibration in~\autoref{sec:frame_alignment}).
The a-priori shared map enables multi-robot coordination and global mission planning, but also provides an additional safety level by allowing robust online analysis of localization drift and cross-checking of sensory measurements.


\subsection{Navigation and Trajectory Tracking}\label{sec:navigation}
The navigation of the \acp{mav} during the mission follows a mission plan $\mathbf{P} \in \mathbb{P}$ generated in the pre-deployment phase, described in~\autoref{sec:predeployment}. 
This collision-free plan is represented by a sequence of triplets \mbox{$\mathbf{P} = \left[\left(\mathbf{p}_{uav}, \mathbf{p}_{ooi}, \mathbb{I}\right)_1,~\dots,~\left(\mathbf{p}_{uav}, \mathbf{p}_{ooi}, \mathbb{I}\right)_{|\mathbf{P}|}\right]$}, where $\mathbb{I} \in \{0, 1\}$ is the acquisition flag.
The triplets with $\mathbb{I} = 1$ specify the \ac{mav} poses $\mathbf{p}_{uav}$ in which capturing an image or illuminating the \ac{ooi} at pose $\mathbf{p}_{ooi}$ is required.
The reference trajectory~$\mathbf{R}$ is generated by uniform sampling of the collision-free path given as sequence of $\mathbf{p}_{uav} \in \mathbf{P}$ such that the sampling step respects the required velocity.
The \ac{mav} is requested to stop at each pose $\mathbf{p}_{uav} \in \mathbf{P}$ where $\mathbb{I} = 1$ to improve the quality of data acquisition by minimizing deviation from the desired pose and reducing the motion blur that would occur in case of non-zero velocity during image capturing.
The reference trajectory~$\mathbf{R}$ then serves as an input to the trajectory tracking module using \ac{mpc}.
This module, described in our previous works~\cite{saska2017etfa, kratky2020ral}, produces a smooth collision-free trajectory while penalizing deviations from the original reference trajectory and respecting dynamic constraints of the \ac{mav}.
The smooth-sampled reference trajectory is then passed into a feedback controller (implemented within the MRS UAV \mbox{system~\cite{baca2021mrs}) handling tracking of the trajectory.}
\subsection{Multi-robot Coordination and Cooperation}\label{sec:multirobot_coordination}

Since the characteristics of the expected environment enable reliable use of standard communication channels,
the cooperation algorithms rely on the information shared through a Wi-Fi interface among the \acp{mav} and a ground station.
Namely, the \acp{mav} share their current poses, planned trajectories, and individual statuses based on the information from their onboard sensors.
The same communication channel is utilized for commanding the \acp{mav} from the ground station in case of emergency or a change in the mission plan, and for sharing specific messages among the \acp{mav} during the realization of cooperative documentation techniques.
The algorithms handling the autonomous flight are computed on board the \acp{mav}.

During the cooperation, the reference trajectories of the \acp{mav} are generated in a distributed manner on a short horizon corresponding to the optimization horizon used in the \acs{mpc}-based trajectory tracking module~\cite{baca2021mrs}.
By applying concepts of leader-follower architectures, the reference trajectories of supporting \acp{mav} are generated with respect to the optimized trajectory of the primary \ac{mav} (leader), to the position and the desired distance of the \ac{mav} from the \ac{ooi}, and to the desired lighting angle with respect to the optical axis of the documentation sensor on board the primary \ac{mav}.
The coordination of the \acp{mav} is part of trajectory optimization (see~\autoref{sec:navigation}) where both the current poses of the \acp{mav} and their planned trajectories are considered to be part of constrained unfeasible \rev{space~}\cite{saska2017etfa}.
\rev{To prevent the downwash effect, this optimization is also constrained to not allow two nearby \mbox{\acp{mav}} to fly above each other.}

\section{Aerial Platforms}\label{sec:aerial_platforms}

\begin{figure*}[htb]
  \centering

  \subfloat[Primary custom-made \ac{mav} application-tailored for documentation and inspection tasks in building interiors.
  The platform carries onboard sensors required for autonomous flight with equipment for acquiring high-quality documentation data, as well as a processing unit for handling autonomous flight, reasoning over the sensory data, obstacle avoidance, and the documentation mission.]
  {\hspace{0.0em}\input{./fig/figure_4a.tex}}\label{fig:naki_mav_description}\\

  \vspace{0.5em}

  \subfloat[Secondary \ac{mav} tailored for supporting documentation tasks in building interiors.
  In contrast to the primary \ac{mav} (a), this platform is smaller and carries a high-power light instead of sensors for the documentation task.]
  {\input{./fig/figure_4b.tex}}\label{fig:supporting_mav_description}
  \hspace{0.2em}
  \subfloat[Comparison of the custom-made \ac{mav} platform (a) with lightweight commercial drone DJI Mavic Air 2, which carries a small camera sensor and does not support complex mission planning in building interiors.]
  {\includegraphics[width=0.99\columnwidth]{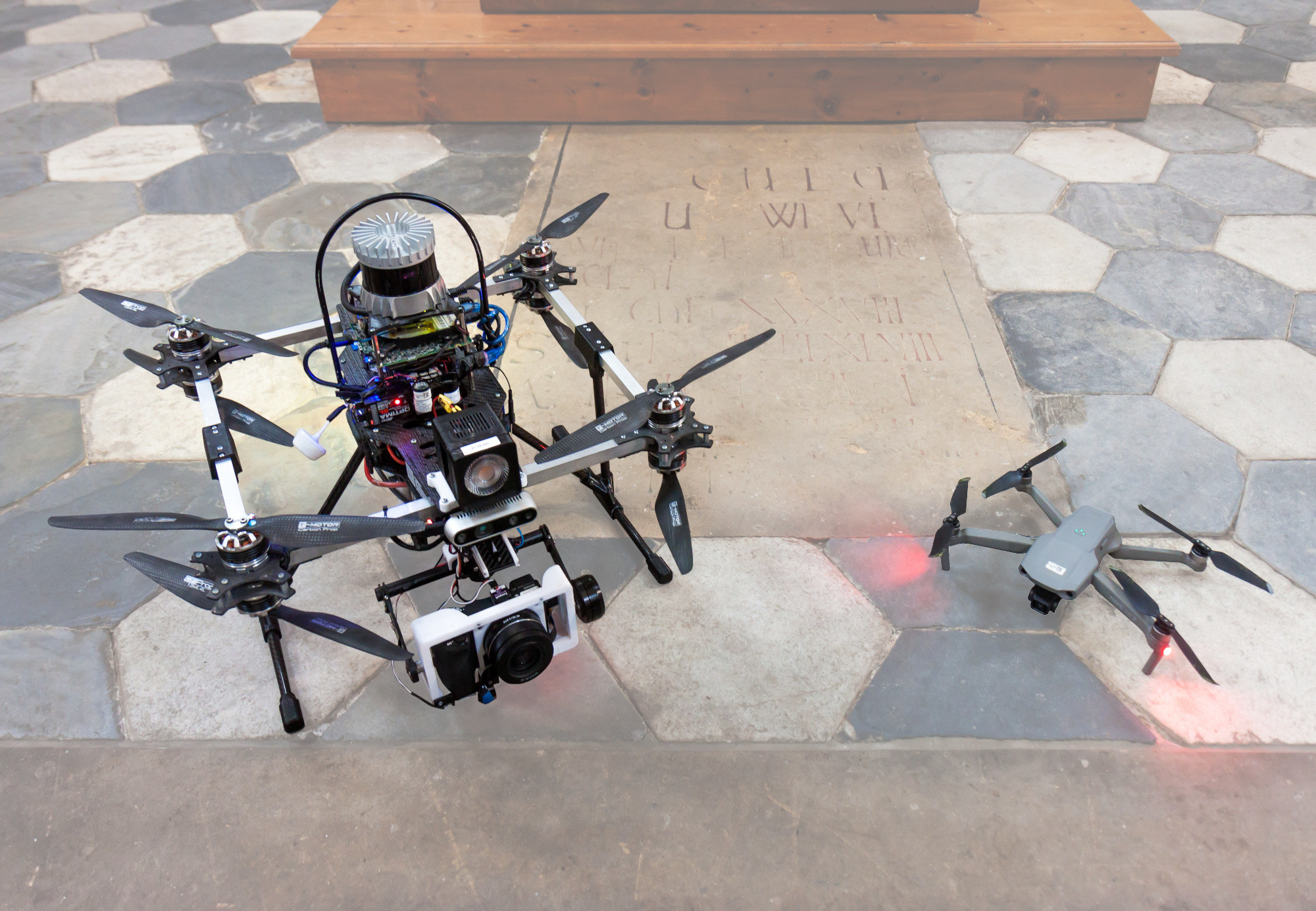}}

  \caption{Aerial platforms used for documentation tasks in the Dronument project --- primary \ac{mav} carrying documentation sensors (a), secondary \ac{mav} assisting in cooperative documentation (b), and commercial drone used for qualitative comparison (c).
  Both (a) and (b) carry environment-perception sensors and computational resources allowing fully autonomous deployment in interiors with poor lighting conditions.}
  \label{fig:hardware}
  \vspace{-0.0em}
\end{figure*}

Two custom-made \ac{mav} platforms were designed specifically for the proposed application of deployment in interiors of buildings.
Both the platforms, as shown in~\autoref{fig:hardware} and described in more detail in~\cite{hert2022hardware}, support fully autonomous deployment within the tackled domain by carrying sensors for local environment perception together with a powerful computational unit handling the entire autonomous aerial mission.
The primary platform is a heavy-weight (\SI{5.5}{\kilo\gram} without payload) octo-rotor with dimensions of \SI[product-units=single]{78x81x40}{\centi\meter}, capable of carrying up to \SI{1.5}{\kilo\gram} payload --- enough for a \ac{mil} camera with a suitable lens and 2-axis gimbal stabilization, as well as an onboard light source.
This platform minimizes its dimensions while maximizing the payload capacities, is equipped with mechanical propeller guards, and carries sensory redundancy for active obstacle avoidance.
The secondary platform is a lightweight (\SI{3}{\kilo\gram} fully loaded) quad-rotor with dimensions of \SI[product-units=single]{68x68x30}{\centi\meter} suited for assisting the primary \ac{mav} throughout a documentation process by providing the scene illumination, thus increasing the quality of the gathered digital materials.
While cooperating, the supporting \acp{mav} assist in performing tasks inexecutable by a single \ac{mav} in principle.
As the primary payload, the secondary platform carries a set of high-power light sources.
Both the platforms support flights in close proximity to obstacles and to other \acp{mav}.
However, relative distances are limited to a minimum of \SI{2}{\meter} to limit the aerodynamic influence of downwash, ceiling, and ground effects on the \ac{mav}, and the contrary effect of the \ac{mav} on the environment (possible damage of not firmly attached objects and fragile plasters).

\subsection{Sensors for Autonomy}

For autonomy in \acs{gnss}-denied environments, both platforms rely on onboard sensors only.
The primary sensor is a 3D \ac{lidar} Ouster OS0-128 with \SI{50}{\meter} detection range and \SI{90}{\degree} vertical \acs{fov} supported by a thermally-stabilized triple-redundancy \acl{imu}, downward and upward looking point-distance sensors Garmin \ac{lidar} Lite, and front-facing (primary \ac{mav}) or downward and upward-facing (secondary \ac{mav}) \acl{rgbd} cameras Intel\textsuperscript{\tiny\textregistered} Realsense D435 for sensory cross-checking in active obstacle avoidance.
All the sensory data are processed by an Intel\textsuperscript{\tiny\textregistered} NUC-i7 onboard computer which utilizes data in real-time algorithms handling the autonomous aerial mission.
The low-level control (attitude stabilization) is handled by Pixhawk~2.1, an open-source autopilot used frequently by the robotic community.
For safety reasons, the primary \ac{mav} carries a visible diagnostic RGB LED which indicates a possible failure to an operator who is authorized to override \ac{mav} autonomy for manual landing.

\subsection{Payload}

The payload equipment mountable on board the platforms is modular --- cameras, lenses, and light sources can be easily interchanged for the purposes of a specific task.
For general purposes, the primary \ac{mav} carries a 2-axis gimbal FlyDrotec capable of stabilizing up to \SI{850}{\gram} payload.
The stabilized axes are controllable, a feature useful mainly for controlling the pitch angle of a camera.
Throughout our experiments, a \ac{mil} camera, the Sony Alpha A6500 with varying lenses, has been used for its integrated image-sensor stabilization, further minimizing the negative effect of mid-flight vibrations on the output image quality.
Triggering image capture is automated via the onboard computer, whereas real-time imaging is transmitted to the ground for online visualization for the operator.




\section{Experiments and Results}\label{sec:exp}

\begin{table*}
  \newcommand{\astable}{1.0}
  \newcommand{\asobject}{1.0}
  \newcommand{\sptop}{\hspace{-0.2cm}}
  \newcommand{\spc}{\phantom{x}}
  \newcommand{\spcp}{\phantom{xi}}
  \newcommand{\spcc}{\phantom{xxx}}
  \newlength{\extravspace}
  \setlength{\extravspace}{1.8ex}
  \setlength\tabcolsep{5.5pt}
  \renewcommand{\arraystretch}{\astable}
  \begin{center}
    \caption{Overview of three-phase \ac{mav} deployment in historical buildings within the Dronument project.
    The first phase focused on specifying the use cases, developing the methodology, designing the system, and performing preliminary experiments, including manually controlled flights.
    The second phase investigated autonomous multi-robot coordination in cooperative documentation and experimented with imaging outside the visible spectrum and with the physical interaction of \acp{mav} with the environment.
    The third phase deployed the system in a full-operation mode for gathering data valuable to end users and for validating the methodology and overall performance of the autonomy.}
    \label{tab:documentation_summary}
    \begin{tabular}{l r r r r r r c c c}
      \toprule
      \vspace{-2em}\\
      \hspace*{-0.2cm}\begin{tabular}{l}\textbf{Object} \rev{(approximate floor area over} \\ \rev{which the system operated)}\end{tabular} &\rot{\sptop flights} & \rot{\sptop images} & \rot{\sptop \begin{tabular}{l} \hspace{-0.2cm}flight time\\(h:mm:ss)\end{tabular}} & \rot{\sptop \begin{tabular}{l} \hspace{-0.2cm}flight \\ dist. (m)\end{tabular}} & \rot{\sptop \begin{tabular}{l} \hspace{-0.2cm}maximum \\ height (m)\end{tabular}} & \rot{\sptop \begin{tabular}{l} \hspace{-0.2cm}min. obst. \\ dist. (m)\end{tabular}} &\rot{\sptop multi-robot} & \rot{\sptop \begin{tabular}{l} \hspace{-0.2cm}applied \\ methods\end{tabular}} & \phantom{xxxxx} \\
      \midrule
      \objecttab{\asobject}{\astable}{\textbf{A}rchbishop's Chateau}{in Krom\v{e}\v{r}\'{i}\v{z} (UNESCO, \rev{420 $\text{m}^2)$}} &10\spc  & 202\spc  & 0:24:30\spcp  & 320\spcc  & 9.0\spcc & 2.0\spcc  & \xmark  & \ac{vis}\spc  & \parbox[t]{0mm}{\multirow{5}{*}[-0.7cm]{\hspace{-0.1cm}\rotatebox[origin=c]{90}{\begin{tabular}{c}\textbf{Phase 1}\\\textbf{2017--2019}\end{tabular}}}} \\[\extravspace]
        \objecttab{\asobject}{\astable}{\textbf{V}ranov nad Dyj\'{i}}{State Chateau \rev{(410 m$^2$)}} & 18\spc  & 1049\spc  & 1:26:44\spcp & 1020\spcc & 12.0\spcc & 3.0\spcc  & \xmark  & \ac{vis}\spc  & \\[\extravspace]
      \objecttab{\asobject}{\astable}{\textbf{K}lein Family Mausoleum}{in Sobot\'{i}n \rev{(30 m$^2$)}} & 7\spc  & 274\spc  & 0:17:37\spcp  & 120\spcc  & 3.8\spcc  & 1.6\spcc  & \xmark  & \ac{vis}\spc  &  \\[\extravspace]
      \objecttab{\asobject}{\astable}{\textbf{R}ondel at State Chateau and Castle}{Jind\v{r}ich\r{u}v Hradec \rev{(140 m$^2$)}} & 8\spc  & 660\spc  & 0:51:32\spcp  & 940\spcc  & 7.8\spcc  & 2.1\spcc  & \xmark  & \ac{vis}\spc  & \\[\extravspace]
      \objecttab{\asobject}{\astable}{\textbf{C}hapel of All Saints at Chateau}{Tel\v{c} (UNESCO\rev{, 84 m$^2$)}} & 6\spc  & 190\spc  & 0:14:04\spcp  & 145\spcc  & 4.8\spcc  & 1.7\spcc  & \xmark  & \ac{vis}\spc  & \\[\extravspace]
      \arrayrulecolor{black!30}
      \objecttab{\asobject}{\astable}{\textbf{C}hurch of St. Mary Magdalene}{in Chlum\'{i}n \rev{(224 m$^2$)}} & 8\spc  & 86\spc  & 0:23:40\spcp  & 146\spcc  & 4.8\spcc  & 1.5\spcc  & \greencheck\spc  & \ac{rti}\spc  & \\[\extravspace]
      \objecttab{\asobject}{\astable}{\textbf{C}hurch of the Holy Trinity}{in B\v{e}ha\v{r}ovice \rev{(252 m$^2$)}} & 6\spc  & 56\spc  & 0:07:10\spcp  & 30\spcc  & 5.2\spcc  & 2.2\spcc  & \xmark  & \ac{irf}, \ac{uvf}, \ac{irr}\spc  & \parbox[t]{0mm}{\multirow{4}{*}[0.35cm]{\hspace{-0.1cm}\rotatebox[origin=c]{90}{\begin{tabular}{c}\textbf{Phase 2}\\\textbf{2019--2021}\end{tabular}}}} \\[\extravspace]
        \objecttab{\asobject}{\astable}{\textbf{C}hurch of St. Maurice}{in Olomouc \rev{(1160 m$^2$)}} & 27\spc  & 971\spc  & 1:31:38\spcp  & 1340\spcc  & 16.8\spcc  & 1.5\spcc  & \greencheck\spc  & \ac{vis}, \ac{tpl}, \ac{rak}\spc  & \\[\extravspace]
      \objecttab{\asobject}{\astable}{\textbf{C}hurch of St. Anne and St. Jacob}{the Great in Star\'{a} Voda \rev{(505 m$^2$)}} & 95\spc  & 6022\spc  & 4:54:10\spcp  & 4540\spcc  & 19.5\spcc  & 1.7\spcc  & \greencheck\spc  & \begin{tabular}{@{}c@{}}\ac{vis}, \ac{tpl}, \ac{rti} \\ \ac{rak}, \ac{uvr}, \ac{irr} \end{tabular}\spc  & \\[\extravspace]
        \objecttab{\asobject}{\astable}{\textbf{C}hurch of the Exaltation of the Holy}{Cross in Prost\v{e}jov \rev{(570 m$^2$)}} & 7\spc  & 548\spc  & 0:19:49\spcp  & 308\spcc  & 15.2\spcc  & 1.7\spcc  & \xmark  & \ac{vis}\spc  & \parbox[t]{0mm}{\multirow{6}{*}[-0.27cm]{\hspace{-0.1cm}\rotatebox[origin=c]{90}{\begin{tabular}{c}\textbf{Phase 3}\\\textbf{2021--2022}\end{tabular}}}}\\[\extravspace]
          \objecttab{\asobject}{\astable}{\textbf{C}hurch of Our Lady of the Snows}{in Olomouc \rev{(918 $\text{m}^2$)}} & 3\spc  & 255\spc  & 0:10:06\spcp  & 185\spcc  & 17.1\spcc  & 2.4\spcc  & \xmark  & \ac{vis}\spc  & \\[\extravspace]
      \objecttab{\asobject}{\astable}{\textbf{C}hurch of the Assumption of the}{Virgin Mary in Cholina \rev{(409 m$^2$)}} & 3\spc  & 82\spc  & 0:08:51\spcp  & 132\spcc  & 7.5\spcc  & 1.4\spcc  & \xmark  & \ac{vis}\spc  & \\[\extravspace]
      \objecttab{\asobject}{\astable}{\textbf{C}hurch of the Nativity of the Virgin}{Mary in Nov\'{y} Mal\'{i}n \rev{(282 m$^2$)}} & 2\spc  & 129\spc  & 0:06:06\spcp  & 68\spcc  & 8.8\spcc  & 1.8\spcc  & \xmark  & \ac{vis}\spc  & \\[\extravspace]
      \objecttab{\asobject}{\astable}{\textbf{C}hurch of the Holy Trinity}{in Kop\v{r}ivn\'{a} \rev{(367 m$^2$)}} & 4\spc  & 211\spc  & 0:17:50\spcp  & 247\spcc  & 11.4\spcc  & 2.0\spcc  & \xmark  & \ac{vis}\spc  & \\[\extravspace]
      \objecttab{\asobject}{\astable}{\textbf{C}hurch of St. Bartholomew}{in Z\'{a}b\v{r}eh \rev{(616 m$^2$)}} & 4\spc  & 263\spc  & 0:18:23\spcp  & 258\spcc  & 12.4\spcc  & 1.4\spcc  & \xmark  & \ac{vis}\spc  & \\
      \arrayrulecolor{black}
      \midrule
      \textbf{Total} \rev{(6387 m$^2$)} & 208\spc  & 10998\spc  & 11:32:10\spcp  & 9799\spcc  & 19.5\spcc  & 1.4\spcc  & \greencheck\spc  & \begin{tabular}{@{}c@{}}\ac{vis}, \ac{tpl}, \ac{rti} \\ \ac{rak}, \ac{uvr}, \ac{irr} \\ \ac{uvf}, \ac{irf} \end{tabular}\spc  & \\
      \bottomrule
    \end{tabular}
  \end{center}

  \begin{tikzpicture}[overlay]
    \draw [gray!80, line width=0.4mm] (17.05, 11.9) to (17.05, 8.7);
    \draw [gray!80, line width=0.4mm] (17.05, 8.4) to (17.05, 5.6);
    \draw [gray!80, line width=0.4mm] (17.2, 7.0) to (17.2, 1.6);
  \end{tikzpicture}

  \vspace{-3mm}

\end{table*}

The extreme requirements on safety imposed by the nature of the application requiring the deployment of \acp{mav} in priceless historical buildings imply thorough validation of all the developed software and hardware solutions prior to their deployment in real-world missions.    
The software solutions ranging from the state estimation and control algorithms to high-level mission control were intensively tested with the use of Gazebo simulator and the MRS simulation package\footnotemark~providing realistic behavior of the \acp{mav}. 
\footnotetext{\href{https://github.com/ctu-mrs/simulation}{github.com/ctu-mrs/simulation}}
Running the same software \rev{with identical parametrization} in simulation and on real hardware significantly simplifies the transfer of algorithms from the virtual environment to real-world applications.
The 3D models built from the obtained 3D scans are directly used as the simulation environments for algorithms' testing.
Together with simulated sensory noises and model inaccuracies, this makes the simulation as analogous to real-world conditions as possible.
This methodology proves to be especially useful for discovering possible failures correlated with specific environments and validating the entire autonomous missions in an approximate copy of the real-world scenarios.
Although the simulator is highly realistic, running the system in the real world introduces additional constraints.
Therefore, even after thorough testing in virtual environments, the first deployments of the system were preceded by test flights in mock-up scenarios and testing interiors in order to reveal potential problems related to transfer of the system from simulation to real hardware. 

The final version of the system, as presented in this manuscript, builds on preliminary versions and architectures of both software and hardware stacks and integrates experience from over a year and a half period of experimental deployments.
During the experimental campaigns, remaining sources of potential failures were identified and the \ac{mav} system upgraded to reach the desired performance and reliability while increasing the number of realizable documentation techniques.
The entire system was, to this day, deployed in real-world missions in fifteen historical buildings of various characteristics (summarized in~\autoref{tab:documentation_summary}), including one of the largest Baroque halls in the Czech Republic at State Chateau Vranov nad Dyj\'{i} and the \mbox{UNESCO} World Heritage Sites, Archbishop's Chateau in Krom\v{e}\v{r}\'{i}\v{z} and Chateau Tel\v{c}.
Almost twelve airborne hours in more than two hundred flights have been performed for purposes of documentation missions in the given structures.
Such an extensive experimental campaign provides an exhaustive validation of the system in real-world conditions and supports its applicability in \acs{gnss}-denied environments by identifying and overcoming challenges imposed by specific scenarios.
The following sections describe the documentation techniques realized by the system in these structures.
The \acp{ooi} of the presented documentation missions are showcased in~\autoref{fig:deployment_mosaic}.

\subsection{Visible Spectrum Photography}

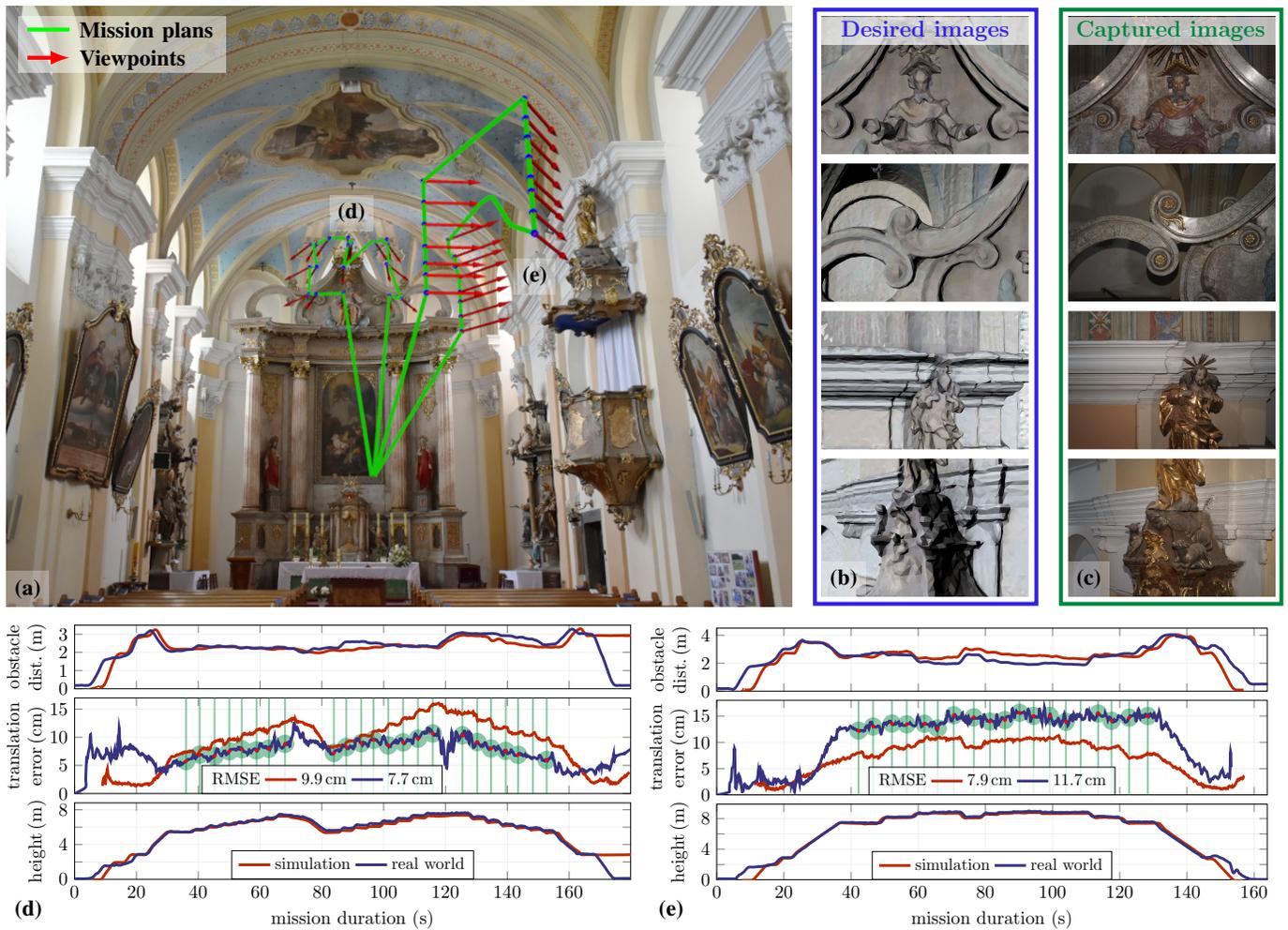
\begin{figure*} [htb]
  \centering

  \newcommand{\xcap}{0.88em}
\newcommand{\xcapb}{1.34em}
\newcommand{\xcapc}{11.23em}
\newcommand{\ycapb}{1.15em}
\newcommand{\ycap}{0.73em}
\newcommand{\fillopa}{0.3}

\begin{tikzpicture}
  \node[anchor=south west,inner sep=0] (a) at (0,0) {\adjincludegraphics[height=0.47\textwidth,trim={{0.15\width} {0.039\height} {0.05\width} {0.043\height}},clip]{\FIGPATH/figure_5a}};%
  \begin{scope}[x={(a.south east)},y={(a.north west)}]
    \node[fill=white, fill opacity=\fillopa, text=black, text opacity=1.0] at (\xcap, \ycap) {\textbf{\small(a)}};
  \end{scope}
  \coordinate (size) at (2.94, 0.795);
  \coordinate (X) at (1.65, 7.99);
  \path[style=very thick,draw=black,draw opacity=0.0,fill=white,fill opacity=0.5] ($(X)-0.5*(size)$) rectangle ($(X)+0.5*(size)$);

  \draw[green,ultra thick] (0.305, 8.2) -- (0.90, 8.2);
  \node[text=black,draw=none,fill=none,anchor=west] at (0.92, 8.16) {\small\textbf{Mission plans}};

  \draw[->, red, ultra thick, -latex] (0.305, 7.80) -- (0.90, 7.80);
  \node[text=black,draw=none,fill=none,anchor=west] at (0.92, 7.77) {\small\textbf{Viewpoints}};

  \node[fill=white, fill opacity=\fillopa, text=black, text opacity=1.0] at (4.9, 5.6) {\textbf{\small(d)}};

  \node[fill=white, fill opacity=\fillopa, text=black, text opacity=1.0] at (7.5, 4.7) {\textbf{\small(e)}};

\end{tikzpicture}
\hfill
\begin{tikzpicture}
  \node[anchor=south west,inner sep=0] (a) at (0,0) {\adjincludegraphics[height=0.47\textwidth,trim={{0.0\width} {0.001\height} {0.0\width} {0.001\height}},clip]{\FIGPATH/figure_5bc}};%
  \begin{scope}[x={(a.south east)},y={(a.north west)}]
    \node[fill=white, fill opacity=\fillopa, text=black, text opacity=1.0] at (\xcapb, \ycapb) {\textbf{\small(b)}};
  \end{scope}
  \begin{scope}[x={(a.south east)},y={(a.north west)}]
    \node[fill=white, fill opacity=\fillopa, text=black, text opacity=1.0] at (\xcapc, \ycapb) {\textbf{\small(c)}};
  \end{scope}
\end{tikzpicture}
\\
\vspace*{0.1cm}
\begin{tikzpicture}
  \node[anchor=south west,inner sep=0] (a) at (0,0) {\adjincludegraphics[width=0.49\textwidth,trim={{0.015\width} {0.03\height} {0.0\width} {0.0\height}},clip]{\FIGPATH/figure_5d}};%
  \begin{scope}[x={(a.south east)},y={(a.north west)}]
    \node[fill=white, fill opacity=0.0, text=black, text opacity=1.0] at (\xcap, \ycap+0.2) {\textbf{\small(d)}};
  \end{scope}
\end{tikzpicture}
\hfill
\begin{tikzpicture}
  \node[anchor=south west,inner sep=0] (a) at (0,0) {\adjincludegraphics[width=0.49\textwidth,trim={{0.015\width} {0.03\height} {0.0\width} {0.0\height}},clip]{\FIGPATH/figure_5e}};%
  \begin{scope}[x={(a.south east)},y={(a.north west)}]
    \node[fill=black, fill opacity=0.0, text=black, text opacity=1.0] at (\xcap, \ycap+0.2) {\textbf{\small (e)}};
  \end{scope}
\end{tikzpicture}
  \caption{An example documentation mission in the Church of the Nativity of the Virgin Mary in Nov\'{y} Mal\'{i}n.
  The documentation mission was divided into two separate flights (a) focused on documenting the upper part of the altar (d) and baldachin of the pulpit (e).
  When going from the top, the rows in (d) and (e) show the minimal distance from the \ac{mav} frame to an obstacle, the 3D position error with image acquisition times (green vertical lines and red dots), and the height above ground.
  The desired imagery specified in the 3D model and the images captured on board are compared in (b) and (c).
  \rev{The simulation data are averaged from 5 runs each.}}
  \label{fig:novy_malin}
\end{figure*}

Imaging in the visible spectrum is the most frequently applied technique as it includes methods providing the widest range of practical information while being relatively easy to perform.
Within the fifteen historical structures, \acp{ooi} of various characteristics have been imaged by autonomous \acp{mav}.
\rev{%
These \mbox{\acp{ooi}} range from artistic elements, such as paintings, stained-glass windows, mosaics, stuccoes, and murals located in the most upper parts of the main naves, to complex 3D structures, such as window frames and altars up to \mbox{\SI{20}{\meter}} high.}
Additionally, objects may include structural damage, such as crevices, cracks, or fractures.

\begin{figure*}[htb]
  \centering

  \input{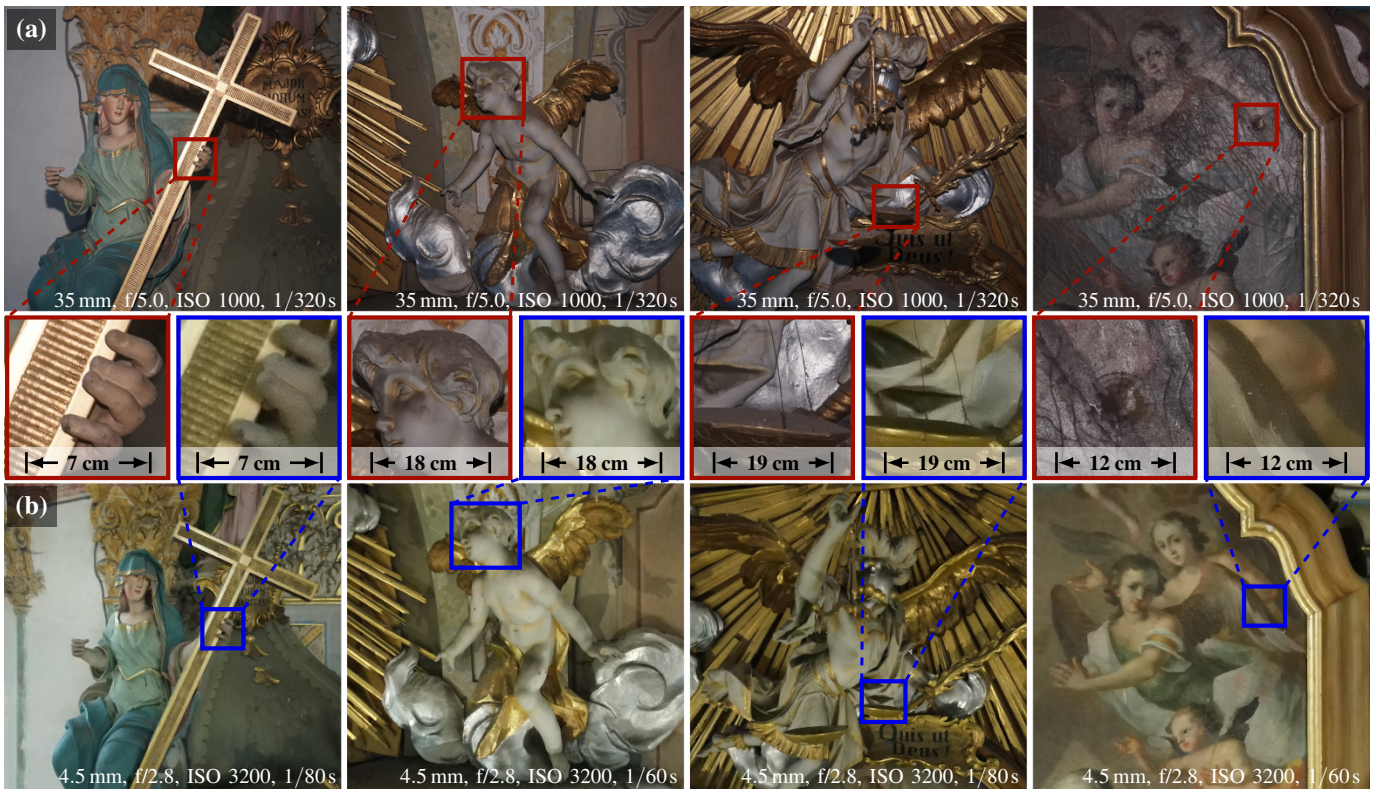}
  \vspace{-1.2em}
  \caption{Image outputs of \ac{vis} methodology as taken by the onboard \ac{mil} camera Sony Alpha A6500 (a) and commercial solution DJI Mavic Air 2 (b).
           Direct comparison of details of the images in the middle row shows that the proposed solution is superior in capturing high-quality details.
           This highlights the last column in which a hole in the painting is visible in top and absent in bottom image.
           Although the commercial solution is small and lightweight, its small sensor size of $\SI[product-units=single]{6.4 x 4.8}{\milli\meter}$ hinders usability in interior documentation.
  }
  \label{fig:dji_comparison}

\end{figure*}

An example of fully-autonomous documentation of a single interior is provided in~\autoref{fig:novy_malin} depicting the documentation of a baroque church.
The specified viewpoints were focused on documentation of two \acp{ooi} --- the upper part of the altar reaching a height of \SI{10}{\meter} and the baldachin of the pulpit.
The automated process of viewpoints' specification and autonomous navigation has enabled fast realization of the documentation process in just two single-\ac{mav} flights lasting only \SI{366}{\second} in total.
With mission specification being part of the pre-deployment phase, the overall time required for in-site deployment reached only \SI{80}{\minute}, including equipment unpacking, flight test, mission validation and execution, and packing.
Such a high level of autonomy in the process demonstrates superiority via fast, safe, effective, and repeatable data capturing when compared to the slow, imprecise, and dangerous manual control of the \ac{mav} in obstacle-filled environments by even a highly trained human operator.
Even with assistive systems (stabilization and collision prevention) guiding the human in navigation, manual operation is unsafe in losses of line of sight in the presence of obstacles and inefficient in time and accuracy required to reach the desired \rev{viewpoints.}
Apart from higher efficiency and safety of autonomy in contrast to human-controlled flying, a fully autonomous system allows flight in close proximity to obstacles, enlarging the operational space of the \ac{mav}.
This is advantageous particularly when documenting elevated \acp{ooi} where the inaccuracy in estimating the \ac{mav}'s distance to the ceiling is proportional to the distance from the human eye, thus making manual navigation in these areas unsafe.

The \ac{vis} method can be performed with commercially available products (e.g., DJI Mavic) offering semi-autonomous solutions in small and lightweight packages.
However, the limited level of autonomy and sensory modularity makes the realization of the missions in large interiors prolonged \rev{(the proposed system is on average ten times faster in the same task)}, non-repeatable, or even impossible in conditions unfavorable to onboard perception or the desired documentation technique.
In~\autoref{fig:dji_comparison}, the images obtained by the proposed system are qualitatively compared to the ones obtained with a commercial product DJI Mavic Air~2.
The figure highlights the superior performance of \ac{mil} camera imaging allowing for capturing high-resolution details of the \acp{ooi} while maintaining a safer distance from the obstacles.

\begin{figure*}
    \centering
    \input{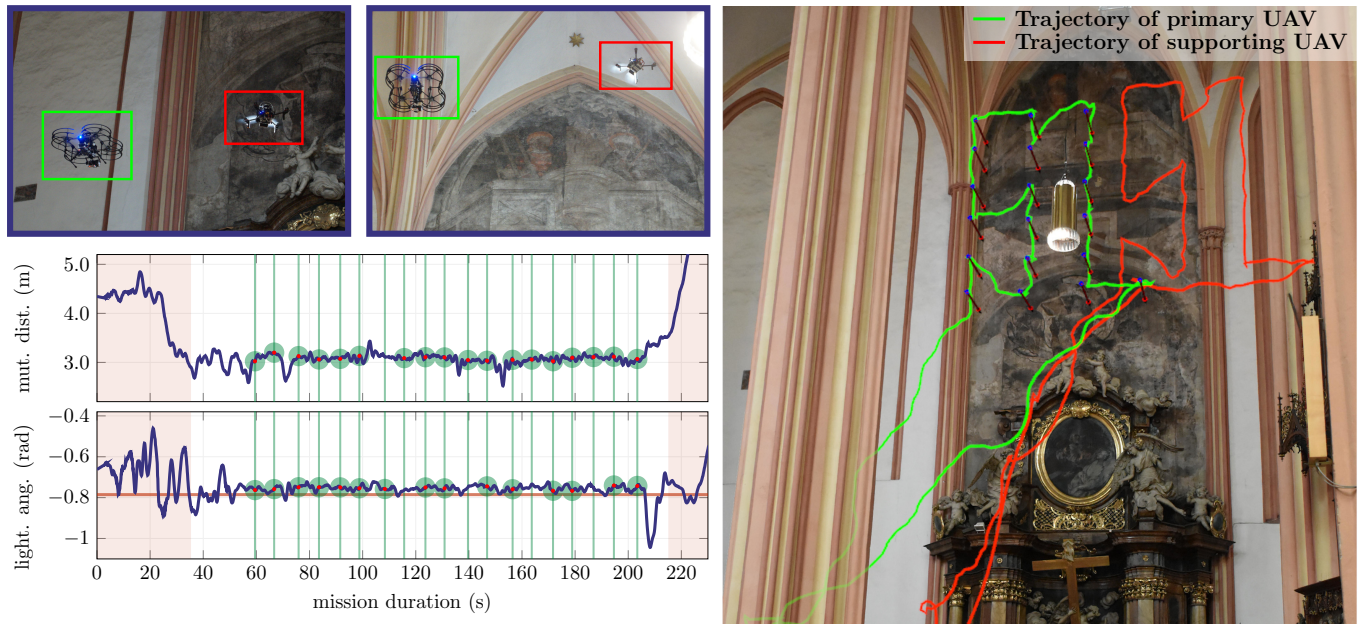}
    \caption{Deployment of a multi-robot formation for detailed documentation of the late Gothic mural of St. Christopher using additional lighting for enhancing the quality of gathered data.
             The graphs show mutual distance of the \acp{mav} during the cooperative flight and the angle between the camera optical axis and light, together with the time occasions of image capturing (green lines and circles).
             The red horizontal line denotes the required angle of lighting.
             The red areas mark parts of the mission in which the \acp{mav} are not required to maintain the formation.%
}
    \label{fig:multirobot_moric}
\end{figure*}

Although \ac{vis} realized by a single \ac{mav} is a powerful technique, a multi-robot approach is often unavoidable if the lighting conditions are insufficient or documentation of an \ac{ooi} requires non-direct lighting.
An example \ac{ooi} requiring additional lighting is the mural of St. Christopher in the late Gothic Church of St. Maurice in Olomouc, the documentation of which is shown in~\autoref{fig:multirobot_moric}.
Insufficient external lighting on the mural did not allow capturing bright, high-quality images without the motion blur effect arising from deviations in the reference pose over a long exposure time.
Thus, to improve the quality of the images, a secondary \ac{mav} provides side lighting (approximately \SI{45}{\degree} with respect to the camera optical axis), lowering exposure times and highlighting details on the mural, such as small crevices invisible to the human eye from the ground.
In the same church, 23 stained-glass windows (each about \SIrange[range-phrase=--,range-units=single]{8}{34}{\meter\squared} large) were able to be documented with a single \ac{mav} as the windows were well illuminated by the outdoor light and could be captured with short exposure times without additional lighting.
The individual images of the mural and the stained-glass windows were rectified and stitched together to compose singular high-resolution orthophotos of each object.
The orthophotos were used to assess the state of the \acp{ooi} for subsequent restoration works and for enhancing the texture of the 3D model of the church\footnotemark.
As compared well in~\cite{kratky2021documentation}, the aerial-based orthophotos outperform the ground-based orthophotos in terms of quality of detail, quality of rectification due to perpendicular optical angles, and absence of occlusions.

\footnotetext{Selected \acsp{ooi} and mapping and 3D reconstruction examples of documented historical structures can be found at \href{http://mrs.felk.cvut.cz/3d-model-viewer/}{mrs.felk.cvut.cz/3d-model-viewer}.}

\subsection{Reflectance Transformation Imaging}
\label{sec:rti}

\ac{rti} method requires a static camera and a dynamic light with a known history of poses.
To validate whether the proposed system is feasible for \ac{rti}, it was applied to document a vault located \SI{11}{\meter} above ground in St. Anne and St. Jacob the Great Church in Star\'{a} Voda (see~\autoref{fig:deployment_mosaic}b). 
This \ac{ooi} was specifically selected as it can be photographed from a balcony on the opposite side of the central nave, thus allowing for the realization of the \ac{rti} technique in two comparable configurations: 1) with the camera (with telephoto lens) mounted on a static tripod with a clear, but misaligned view on the vault and 2) with the camera mounted on board the primary \ac{mav}.
In both configurations, the light was carried on board the secondary \ac{mav}, with the directions of illumination being derived from the poses of this \ac{mav}, as estimated on board during the flight.

\begin{figure*}
  \centering
  \input{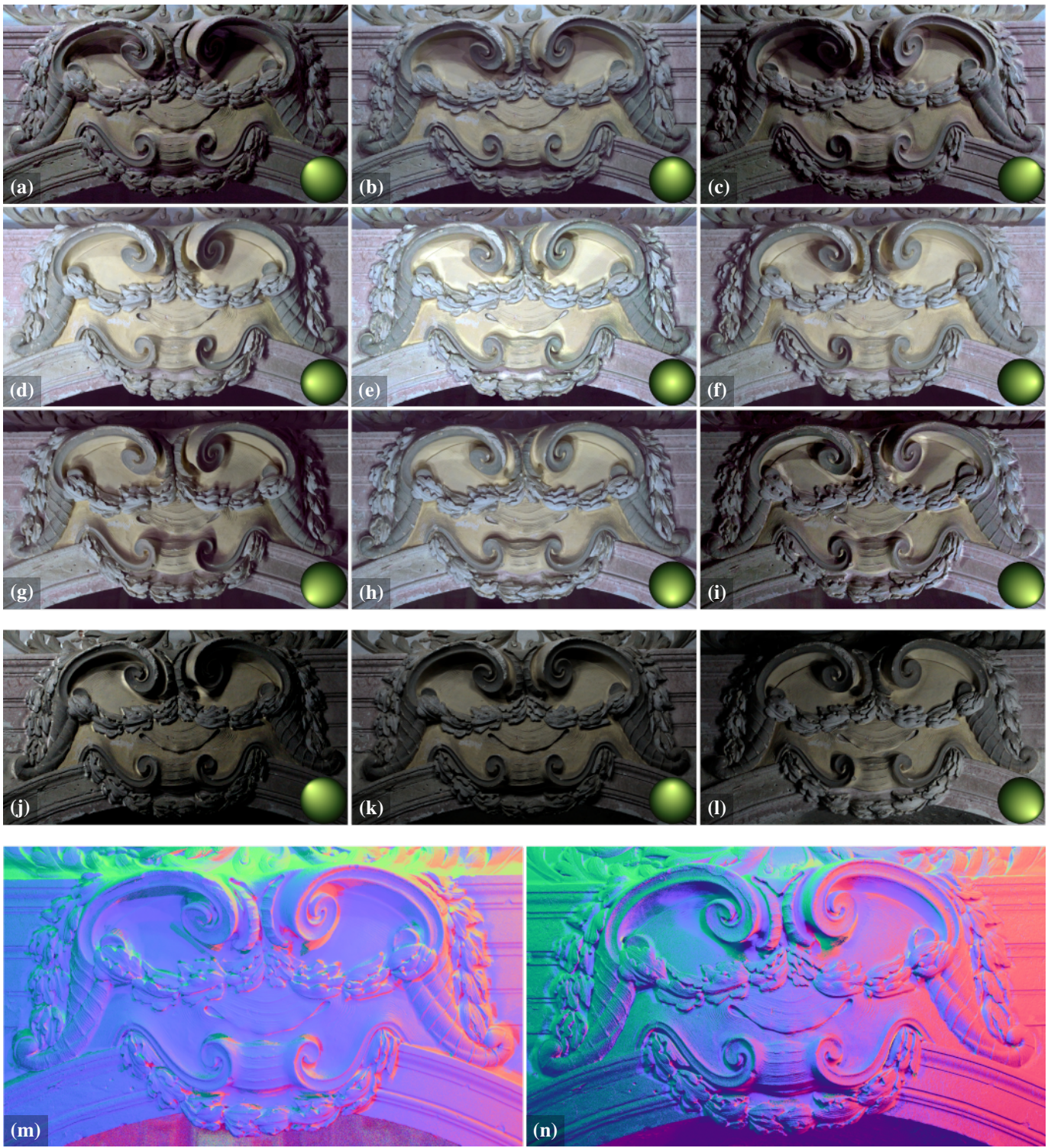}
  \vspace{-1.2em}
  \caption{Comparison of \aclp{ptm} (\acs{ptm}) obtained with a fully \acs{mav}-based \ac{rti} approach with camera carried by a \ac{mav} (a)--(i) and \ac{ptm} obtained from images taken by a camera mounted on a static tripod (j)--(l).
  In both cases, the dynamic positioning of light is provided by the secondary \ac{mav}.
  The bottom row shows the normal maps encoded in RGB for fully \acs{mav}-based approach (m) and a single \ac{mav} approach (n).}
  \label{fig:rti_comparison}
\end{figure*}

The comparison of results obtained in each configuration is presented in~\autoref{fig:rti_comparison}.
The image representation produced from images captured by the tripod-mounted camera yields higher quality, as the choice of the \ac{ooi} and usage of a telephoto lens fully compensate for the main disadvantages of the methodology in this particular case.
These disadvantages are primarily smaller operational space, lower detail resolution of the resulting image caused by the large distance of the camera from the \ac{ooi}, and often unavoidable occlusions.
Although the fully \acs{mav}-based approach yields lower image quality since the camera's pose is not static over time, it has wider operational space and enables imaging from appropriate angles, as was verified for other \acp{ooi} in the church that could not be reasonably captured by a static camera at all.
The non-staticity of the camera's reference pose misaligns the images; thus, their sub-pixel post-alignment is required to avoid blur in the resulting \ac{ptm}.
The experiment shows that the fully \acs{mav}-based approach yields comparable results to the single-\ac{mav} approach, which is favorable when the \ac{ooi} can be photographed from the ground --- an impossible scenario for most \acp{ooi} in difficult-to-reach areas of historical buildings.
  


\subsection{Raking Light and Environmental Monitoring}

A common feature of raking light documentation and monitoring of environmental conditions with \acp{mav} stands in the need for robot-environment interaction.
In the former, a light is attached to the wall illuminating a planar \ac{ooi} from a direction perpendicular to the optical axis of the camera.
This method is known to highlight even the smallest crevices and cracks in the planar surface.
For the latter, a wireless sensor (e.g., for measuring humidity or temperature) is attached to the wall to measure the environmental conditions over longer periods of time.
For the purpose of physical environment-\ac{mav} interaction itself, we researched a \ac{mav} equipped with a system for admittance-based control allowing for stabilization while being attached to a planar surface (and possibly interacting with it)~\cite{smrcka2021icuas}.
Before using this technology, the involved risks must be compared to the payoff, particularly inside historical buildings.
To minimize the risks, it is more convenient to interact with structural (not artistic) parts of the buildings.
The system was successfully tested in real-world mock-up scenarios (see~\autoref{fig:mav_other_techniques}c) with walls of sufficiently good condition.

\subsection{IR and UV Photography}

Realization of \ac{uvf} and \ac{irf} (fluorescent photography) is methodically similar to \ac{vis} with the equipment being a standard \ac{mil} camera and a source of light at appropriate frequency.
In contrast to \ac{vis}, the light emitted by the object illuminated by an IR or UV light source in the visible spectrum is lower.
Thus, these methods require higher exposure times, as specified in~\autoref{tab:exposure_times}.
The higher exposure times put stricter requirements on image stabilization in the presence of onboard vibrations, inaccuracies, and disturbances that cause \acp{mav} to deviate from their reference pose.

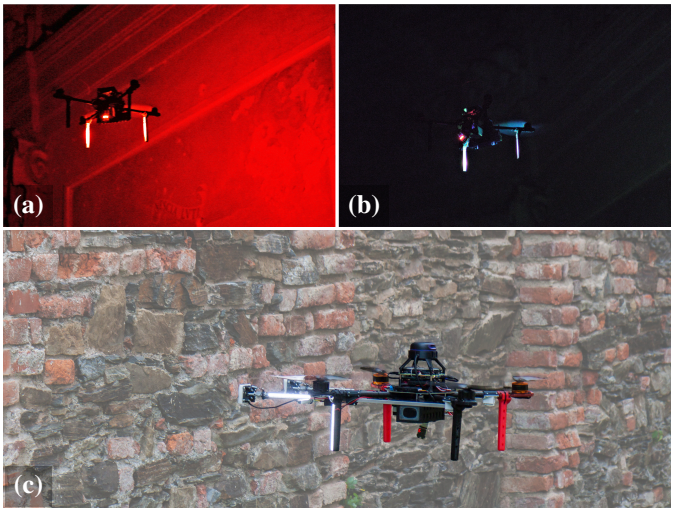
\begin{figure} [t]
  \centering
  \newcommand{\xcap}{0.95em}
\newcommand{\xcapb}{13.65em}
\newcommand{\ycap}{0.8em}
\newcommand{\fillopa}{0.3}

\begin{tikzpicture}
  \node[anchor=south west,inner sep=0] (a) at (0,0) {\adjincludegraphics[width=1.0\columnwidth,trim={{0.00\width} {0.0\height} {0.00\width} {0.00\height}},clip]{\FIGPATH/figure_9ab_bckg}};
  \begin{scope}[x={(a.south east)},y={(a.north west)}]
    \node[fill=black, fill opacity=\fillopa, text=white, text opacity=1.0] at (\xcap, \ycap) {\textbf{(a)}};
    \node[fill=black, fill opacity=\fillopa, text=white, text opacity=1.0] at (\xcapb, \ycap) {\textbf{(b)}};
  \end{scope}
\end{tikzpicture}

\vspace*{0.05cm}

\begin{tikzpicture}
  \node[anchor=south west,inner sep=0] (c) at (0,0) {\adjincludegraphics[width=1.0\columnwidth,trim={{0.00\width} {0.0\height} {0.00\width} {0.00\height}},clip]{\FIGPATH/figure_9c_bckg}};
  \begin{scope}[x={(b.south east)},y={(b.north west)}]
    \node[fill=black, fill opacity=\fillopa, text=white, text opacity=1.0] at (\xcap, \ycap) {\textbf{(c)}};
  \end{scope}
\end{tikzpicture}
  \caption{Deployment of \acp{mav} carrying \ac{ir} (a) and \ac{uv} (b) source of light, and a frame-extension mechanism for physical attachment and interaction with static planar surfaces (c).}
  \label{fig:mav_other_techniques}
  \vspace{-3mm}
\end{figure}

Realization of the \ac{uv} and \ac{ir} reflectography requires a camera without UV and IR filters and exposure times of tens of seconds.
This makes the use of \acp{mav} for imaging in \ac{uv} and \ac{ir} reflectography unfeasible.
However, supporting ground-based imaging with aerial lighting is applicable. 
The \acp{mav} can carry (relatively close to the \ac{ooi}) high-power LEDs radiating in the desired spectrum.
The \ac{ir} and \acs{uv}-based methods were tested in St. Anne and St. Jacob the Great Church, Star\'{a} Voda (see~\autoref{fig:mav_other_techniques}) and in Church of the Holy Trinity, B\v{e}ha\v{r}ovice.
The experiments showed that the proposed system can be used in realization of the \ac{uv} and \ac{ir}-based methods in historical structures, even in limited lighting conditions.

\subsection{Mapping and 3D Reconstruction}

\begin{figure}[t]
  \centering
  \input{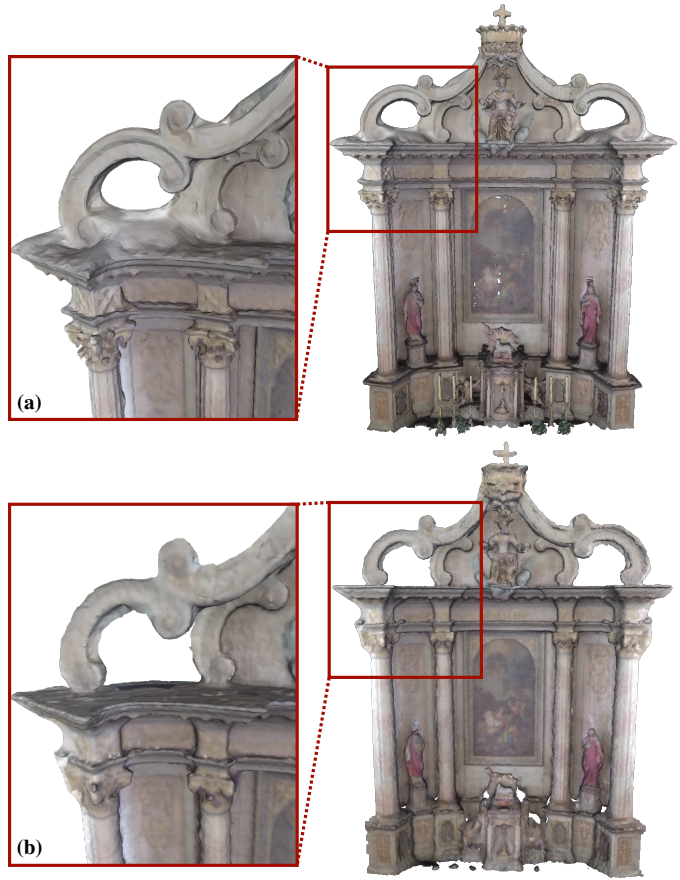}
  \vspace{-1.2em}
  \caption{3D reconstruction of the altar at the Church of the Nativity of the Virgin Mary in Nov\'{y} Mal\'{i}n, Czech Republic.
           The altar reconstructions were done using scans obtained by (a) terrestrial laser scanner Leica BLK360 and (b) Ouster OS0-128 mounted on board an autonomous \ac{mav} during the deployment shown in~\autoref{fig:novy_malin}.
           The meshes were created with the Poisson surface reconstruction and colored using the panoramic RGB images captured by the terrestrial scanner.}
  \label{fig:novy_malin_oltar}
  \vspace{-3mm}
\end{figure}

The capacities of \acp{mav} allow capturing the interior under difficult-to-reach angles, not only for imaging purposes, but also for spatial mapping of the structures.
Although terrestrial laser scanners yield the most accurate maps, these devices cannot, in principle, document occluded spaces, whereas the larger operational space of \acp{mav} allows for minimizing these occlusions.
This advantage is showcased in~\autoref{fig:novy_malin_oltar} where above-ledge areas could not be reconstructed from scans captured at ground.
The potential for accurate 3D mapping using \acp{mav} is immense; however, is not the main purpose of the proposed system which outputs dense 3D maps only as a byproduct to the photo-documentation task.
The onboard-\ac{mav}-built maps contain larger amounts of noise as the mobile laser-scanning technology is less accurate (lightweight, low-power, and moving while scanning) than static scanners, making it harder to align the captured scans, even in post-processing.
To achieve the best results for 3D reconstruction, we recommend leveraging the advantages of both methodologies simultaneously.

\section{Discussion}

The proposed \ac{mav}-based system for documenting historical monuments of differing structures, dimensions, and complexity has demonstrated its wide applicability in real-world documentation tasks, ranging from RGB photography and 3D mapping to multi-robot \acs{rti} in areas high above the ground. 
The high level of autonomy, the ability to fly beyond the visual line of sight between the UAV and a human operator, and the deployability in low lighting conditions (using a worldwide unique method of dynamic illumination by a cooperating UAV team) enable to gather crucial data for heritage protection and documentation that was not possible before.
This universally novel system has been used in the very first fully-autonomous multi-robot real-world deployments in such complex and safety-demanding interior structures.

However, deploying mobile robots inherently poses risks to the environment, humans, and equipment therein.
This requires careful justification of the \acp{mav}' use that, in our experience, tends to be needlessly overused --- conventional technology provides a safer and better quality solution in many documentation tasks.
A common example is imaging the interior ceiling or low-height \acp{ooi}, where using a static camera with a long-focus lens was identified to be a more appropriate solution.
Manual-control \ac{mav} solutions are also sufficient if the task is small-scale, the lighting conditions are feasible, repeatability is not required, and the \acp{ooi} are few.
The need for multi-\ac{mav} teams in tasks achievable with sufficient quality by a single \ac{mav}, such as the selected example of single-\ac{mav} \ac{rti} presented in~\autoref{fig:rti_comparison}, should also be considered prior a full-scale deployment.
\section{Conclusion}\label{sec:clr}

This work has presented a universally novel study on an autonomous multi-robot \ac{mav}-based system for realization of advanced documentation techniques in culturally valuable environments.
The system showcases the immense potential of mobile robots for fast, accurate, and mobile digitalization of difficult-to-access interiors.
The hardware and software architectures of the self-contained autonomous-\acs{mav}-based system were introduced and experimentally validated through almost twelve hours of flight time in more than two hundred real-world flights of single-\acp{mav} and multi-\ac{mav} teams in fifteen historical monuments of varying structures.
The system design has emerged from close cooperation with a team of restorers, and the data collected during the autonomous missions has been used by the end users in successive restoration works.

The study also assists in identifying the current challenges and future directions of research in aerial documentation and inspection.
Based on the high added value for heritage protection, the system has been approved by the Czech National Heritage Institute for indoor usage and is accompanied by an official methodology (available at~\cite{mrs_dronument_page}) describing the proper usage of \acp{mav} in historical structures.
It is the first methodology of this authority for using \acp{mav} in historical buildings and so prescribes the system to be a standard in this application.

\section{Acknowledgment}\label{sec:acknowledgments}
This work was supported
by the Ministry of Culture of the Czech Republic through project no. \href{https://starfos.tacr.cz/en/project/DG18P02OVV069}{DG18P02OVV069} in program NAKI II,
by the Ministry of Education of the Czech Republic through OP VVV funded project CZ.02.1.01/0.0/0.0/16 019/0000765 "Research Center for Informatics,"
by the European Union’s Horizon 2020 research and innovation program AERIAL-CORE under grant agreement no. 871479,
by CTU grant no. SGS20/174/OHK3/3T/13, and
by the Czech Science Foundation (GA\v{C}R) under research project No. 20-10280S.

The authors would like to thank
Nicolas Staub for an initial survey of the projects' applicability in documentation and restoration,
Jan Bednar for his assistance in creating 3D models for public presentation,
Pavel Stoudek for preparing hardware platforms used during the experimental analyses, and
Vojtech Krajicek for technical consultation of the restoration techniques.
At last, we would like to thank to representatives of the Czech National Heritage Institute, namely Milan Skobrtal and Michaela Cadilova, for our fruitful cooperation during the project and for the opportunity to deploy the proposed system in real-world structures.


\bibliographystyle{IEEEtran}
\bibliography{main}

\end{document}